\documentclass[10pt,journal,compsoc]{IEEEtran}

\usepackage{graphicx}
\usepackage{amsmath}
\usepackage{amsthm}
\usepackage{booktabs}
\usepackage{amssymb}
\usepackage{array}
\usepackage{siunitx}
\usepackage{hyperref}
\usepackage{xcolor}
\usepackage{xspace}
\usepackage{longtable}
\usepackage{marvosym}
\usepackage{enumerate}
\usepackage{forest}
\usepackage{mathrsfs}
\usepackage{amsfonts}
\usepackage[numbers]{natbib}

\newtheorem{theorem}{Theorem}
\newtheorem{definition}{Definition}

\newcommand{\dg}{domain generalization\xspace}
\newcommand{\da}{domain adaptation\xspace}

\newcommand{\ieno}{\textit{i}.\textit{e}.}
\newcommand{\etcno}{\textit{etc}} 

\ifCLASSINFOpdf
\else

\fi

\begin{document}
\title{Generalizing to Unseen Domains: A Survey on Domain Generalization}
\author{Jindong~Wang,
        Cuiling~Lan,
        Chang~Liu,
        Yidong~Ouyang,
        Tao~Qin, \IEEEmembership{Senior~Member,~IEEE},
        Wang~Lu,
        Yiqiang~Chen, \IEEEmembership{Senior~Member, ~IEEE},
        Wenjun Zeng,
        \IEEEmembership{Fellow, ~IEEE},
        Philip S. Yu,
        \IEEEmembership{Fellow, ~IEEE}
\IEEEcompsocitemizethanks{\IEEEcompsocthanksitem J. Wang, C. Lan, C. Liu, W. Zeng, and T. Qin are with Microsoft Research Asia, Beijing, China. Correspondence to: Jindong Wang.\protect\\
E-mail: \{jindong.wang,culan,changliu,wezeng,taoqin\}@microsoft.com.
\IEEEcompsocthanksitem Y. Ouyang is with  School of Data Science, Chinese University of Hong Kong, Shenzhen. Email:  yidongouyang@link.cuhk.edu.cn.
\IEEEcompsocthanksitem W. Lu and Y. Chen are with Institute of Computing Technology, Chinese Academy of Sciences, Beijing, China.
\IEEEcompsocthanksitem P. Yu is with University of Illinois at Chicago, and Institute for Data Science, Tsinghua University. Email: psyu@uic.edu.
}

}

\markboth{Accepted by IEEE Transactions on Knowledge and Data Engineering}
{Wang \MakeLowercase{\textit{et al.}}: Generalizing to Unseen Domains: A Survey on Domain Generalization}
\IEEEtitleabstractindextext{%
\begin{abstract}
Machine learning systems generally assume that the training and testing distributions are the same. To this end, a key requirement is to develop models that can generalize to unseen distributions. Domain generalization (DG), \ieno, out-of-distribution generalization, has attracted increasing interests in recent years. Domain generalization deals with a challenging setting where one or several different but related domain(s) are given, and the goal is to learn a model that can generalize to an \emph{unseen} test domain. Great progress has been made in the area of domain generalization for years. This paper presents the first review of recent advances in this area. First, we provide a formal definition of domain generalization and discuss several related fields.
We then thoroughly review the theories related to domain generalization and carefully analyze the theory behind generalization. We categorize recent algorithms into three classes: data manipulation, representation
learning, and learning strategy, and present several popular algorithms in detail for each category. Third, we introduce the commonly used datasets, applications, and our open-sourced codebase for fair evaluation. Finally, we summarize existing literature and present some potential research topics for the future.
\end{abstract}

}

\maketitle
\IEEEdisplaynontitleabstractindextext
\IEEEpeerreviewmaketitle

\section{Introduction}
\label{sec:introduction}

\IEEEPARstart{M}{achine} learning (ML) has achieved remarkable success in various areas, such as computer vision, natural language processing, and healthcare.
The goal of ML is to design a model that can learn general and predictive knowledge from training data, and then apply the model to new (test) data. 
Traditional ML models are trained based on the \emph{i.i.d.} assumption that training and testing data are identically and independently distributed.
However, this assumption does not always hold in reality.
When the probability distributions of training data and testing data are different, the performance of ML models often deteriorates due to domain distribution gaps~\cite{quinonero2009dataset}. 
Collecting the data of all possible domains to train ML models is expensive and even prohibitively impossible. Therefore, enhancing the \emph{generalization} ability of ML models is important in both industry and academic fields.

\begin{figure}[t!]
    \centering
    \includegraphics[width=\columnwidth]{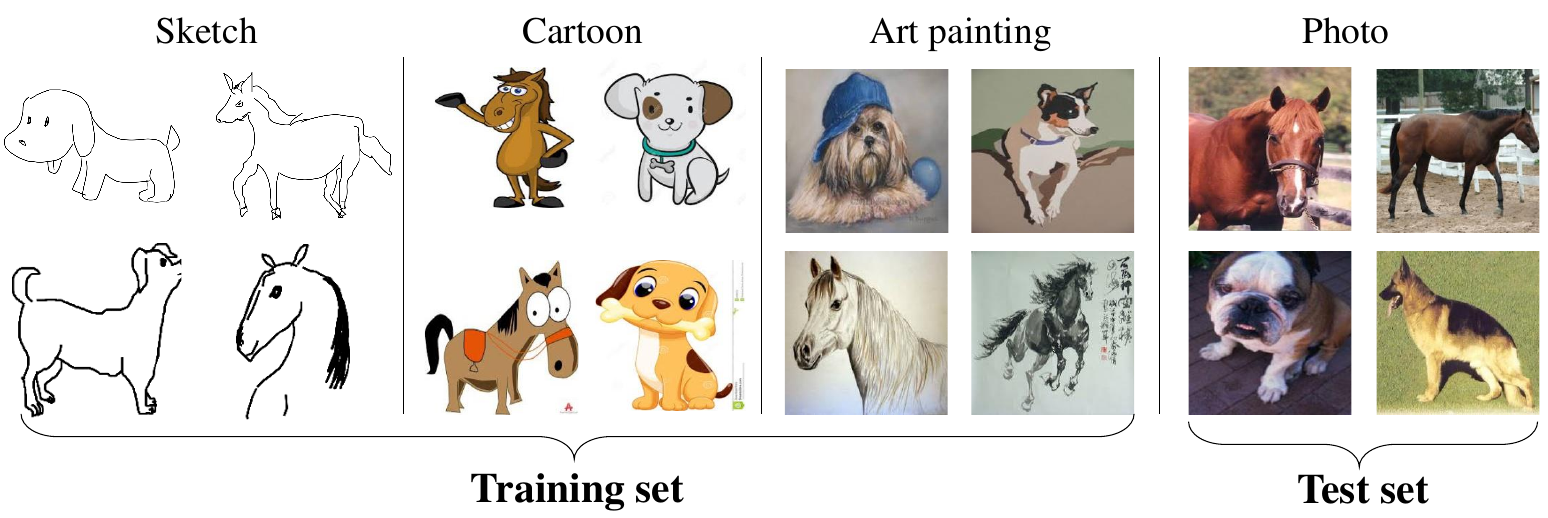}
    \caption{Examples from the dataset PACS~\cite{li2017deeper} for \dg. The training set is composed of images belonging to domains of sketch, cartoon, and art paintings. DG aims to learn a generalized model that performs well on the unseen target domain of photos.}
    \label{fig-dg-example}
    \vspace{-.1in}
\end{figure}

There are many generalization-related research topics such as domain adaptation, meta-learning, transfer learning, covariate shift, and so on.
In recent years, \emph{Domain generalization (DG)} has received much attention.
As shown in \figurename~\ref{fig-dg-example}, the goal of \dg is to learn a model from one or several different but related domains (\ieno, diverse training datasets) that will generalize well on \emph{unseen} testing domains.
For instance, given a training set consisting of images coming from sketches, cartoon images and paintings, domain generalization requires to train a good machine learning model that has minimum prediction error in classifying images coming from natural images or photos, which are clearly having distinguished distributions from the images in training set.
Over the past years, \dg has made significant progress in various areas such as computer vision and natural language processing.
Despite the progress, there has not been a survey in this area that comprehensively introduces and summarizes its main ideas, learning algorithms and other related problems to provide research insights for the future.

In this paper, we present the first survey on \dg to introduce its recent advances, with special focus on its formulations, theories, algorithms, research areas, datasets, applications, and future research directions.
We hope that this survey can provide a comprehensive review for interested researchers and inspire more research in this and related areas.

There are several survey papers after the conference version of our paper, and they are significantly different from ours.
Concurrently, \citet{zhou2021domain} also wrote a survey on DG, while their focus is in computer vision field.
A more recent survey paper is on out-of-distribution (OOD) generalization by \citet{shen2021towards}.
Their work focused on causality and stable neural networks.
A related survey paper~\cite{yang2021generalized} is for OOD detection instead of building a working algorithm that can be applied to any unseen environments.

This paper is a heavily extended version of our previously accepted short paper at IJCAI-21 survey track (6 pages, included in the appendix file).
Compared to the short paper, this version makes the following extensions:
\begin{itemize}
    \item We present the theory analysis on domain generalization and the related domain adaptation.
    \item We substantially extend the methodology by adding new categories: \textit{e.g.,}  causality-inspired methods, generative modeling for feature disentanglement, invariant risk minimization, gradient operation-based methods, and other learning strategies to comprehensively summarize these DG methods.
    \item For all the categories, we broaden the analysis of methods by including more related algorithms, comparisons, and discussion. And we also include more recent papers (over 30\% of new work).
    \item We extend the scope of datasets and applications, and we also explore evaluation standards to domain generalization. Finally, we build an open-sourced codebase for DG research named \emph{DeepDG}\footnote{\url{https://github.com/jindongwang/transferlearning/tree/master/code/DeepDG}} and conduct some analysis of the results on public datasets.
\end{itemize}

This paper is organized as follows.
We formulate the problem of domain generalization and discuss its relationship with existing research areas in Section~\ref{sec-back-related}.
Section~\ref{sec-theory} presents the related theories in \dg.
In Section~\ref{sec-method}, we describe some representative DG methods in detail.
In Section~\ref{sec-research-area}, we show some new DG research areas extended from the traditional setting.
Section~\ref{sec-app} presents the applications and Section~\ref{sec-dataset} introduces the benchmark datasets for DG.
We summarize the insights from existing work and present some possible future directions in Section~\ref{sec-diss}.
Finally, we conclude this paper in Section~\ref{sec-con}.

\section{Background}
\label{sec-back-related}

\subsection{Formalization of Domain Generalization}
In this section, we introduce the notations and definitions used in this paper.

\begin{definition}[Domain]
Let $\mathcal{X}$ denote a nonempty input space and $\mathcal{Y}$ an output space.
A domain is composed of data that are sampled from a distribution. We denote it as $\mathcal{S} = \{(\mathbf{x}_i, y_i)\}^n_{i=1} \sim P_{XY}$, where $\mathbf{x} \in \mathcal{X} \subset \mathbb{R}^d$, $y \in \mathcal{Y} \subset \mathbb{R}$ denotes the label, and $P_{XY}$ denotes the joint distribution of the input sample and output label. $X$ and $Y$ denote the corresponding random variables.
\end{definition}

\begin{figure}[t!]
    \centering
    \includegraphics[width=.48\textwidth]{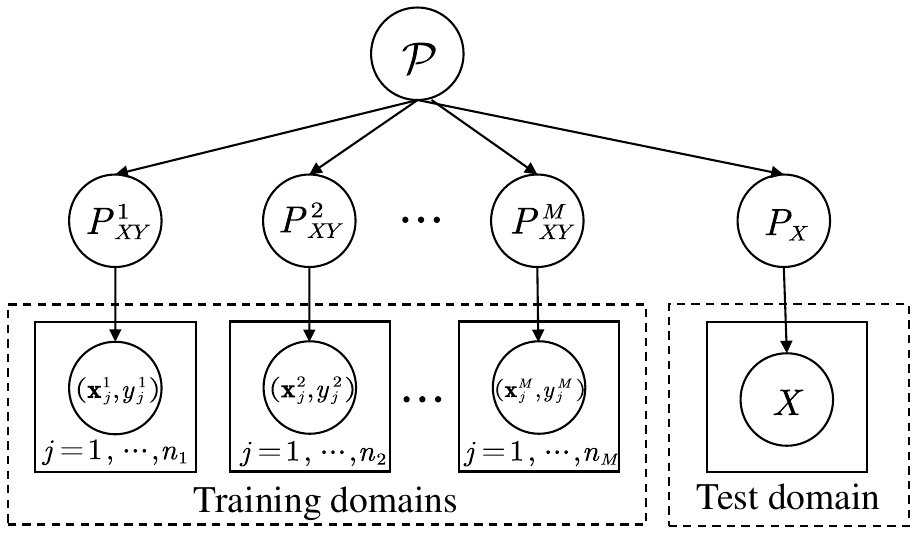}
    \caption{Illustration of \dg. Adapted from \cite{blanchard2011generalizing}.}
    \label{fig-dg-dist}
\end{figure}

\begin{definition}[Domain generalization]
\label{def-dg}
As shown in \figurename~\ref{fig-dg-dist}, in \dg, we are given $M$ training (source) domains $\mathcal{S}_{train}=\{\mathcal{S}^i \mid i=1,\cdots,M\}$ where $\mathcal{S}^i = \{(\mathbf{x}^i_j, y^i_j)\}^{n_i}_{j=1}$ denotes the $i$-th domain.
The joint distributions between each pair of domains are different: $P^i_{XY} \ne P^j_{XY}, 1 \le i \ne j \le M$.
The goal of \dg is to learn a robust and generalizable predictive function $h: \mathcal{X} \to \mathcal{Y}$ from the $M$ training domains to achieve a minimum prediction error on an \emph{unseen} test domain $\mathcal{S}_{test}$ (\ieno, $\mathcal{S}_{test}$ cannot be accessed in training and $P^{test}_{XY} \ne P^i_{XY} \text{ for }i \in \{1,\cdots,M\}$):
\begin{equation}
    \min_{h} \, \mathbb{E}_{(\mathbf{x},y) \in \mathcal{S}_{test}} [ \ell(h(\mathbf{x}),y) ],
\end{equation}
where $\mathbb{E}$ is the expectation and $\ell(\cdot, \cdot)$ is the loss function.
\end{definition}

We list the frequently used notations in \tablename~\ref{tb-notation}.

\begin{table}[htbp]
\caption{Notations used in this paper.}
\label{tb-notation}
\resizebox{.5\textwidth}{!}{
\begin{tabular}{ll|ll}
\toprule
Notation & Description & Notation & Description \\ \hline
$\mathbf{x}, y$ & Instance/label & $\ell(\cdot, \cdot)$ & Loss function \\
$\mathcal{X}, \mathcal{Y}$ & Feature/label space & $h$ & Predictive function \\ 
$\mathcal{S}$ & Domain & $g,f$ & Feature extractor/classifier \\ 
$P(\cdot)$ & Distribution & $\epsilon$ & Error (risk) \\ 
$\mathbb{E}[\cdot]$ & Expectation & $\theta$ & Model parameter \\
$M$ & Number of source domain & $n_i$ & Data size of source domain $i$ \\
\bottomrule
\end{tabular}
}
\end{table}

\subsection{Related Research Areas}

There are several research fields closely related to domain generalization, including but not limited to: transfer learning, domain adaptation, multi-task learning, multiple domain learning, meta-learning, lifelong learning, and zero-shot learning. We summarize their differences with \dg in \tablename~\ref{tb-related} and briefly describe them in the following.\looseness=-1

\textbf{Multi-task learning}~\citep{caruana1997multitask} jointly optimizes models on several related tasks. By sharing representations between these tasks, we could enable the model to generalize better on the original task.  
Note that multi-task learning does not aim to enhance the generalization to a new (unseen) task.
Particularly, multi-domain learning is a kind of multi-task learning, which trains on multiple related domains to learn good models for each original domain~\citep{zhou2020domain} instead of new test domains.

\textbf{Transfer learning}~\citep{Pan2010ASO,weiss2016survey,zhuang2020comprehensive} trains a model on a source task and aims to enhance the performance of the model on a different but related target domain/task.
Pretraining-finetuning is the commonly used strategy for transfer learning where the source and target domains have different tasks and target domain is accessed in training.
In DG, the target domain cannot be accessed and the training and test tasks are often the same while they have different distributions.\looseness=-1

\begin{table*}[t!]
\centering
\caption{Comparison between \dg and some related learning paradigms.}
\label{tb-related}
\begin{tabular}{llllc}
\toprule
Learning paradigm & Training data & Test data & Condition & Test access\\ \hline
Multi-task learning & $\mathcal{S}^1, \cdots, \mathcal{S}^n$ & $\mathcal{S}^1, \cdots, \mathcal{S}^n$ & $\mathcal{Y}^i \ne \mathcal{Y}^j, 1 \le i \ne j \le n$ & $\checkmark$\\ 
Transfer learning & $\mathcal{S}^{src}$, $\mathcal{S}^{tar}$ & $\mathcal{S}^{tar}$ & $\mathcal{Y}^{src} \ne \mathcal{Y}^{tar}$ &  $\checkmark$ \\ 
Domain adaptation & $\mathcal{S}^{src}, \mathcal{S}^{tar}$ & $\mathcal{S}^{tar}$ & $P(\mathcal{X}^{src}) \ne P(\mathcal{X}^{tar})$ &  $\checkmark$\\ 
Meta-learning & $\mathcal{S}^1, \cdots, \mathcal{S}^n$ & $\mathcal{S}^{n+1}$ & $\mathcal{Y}^i \ne \mathcal{Y}^j, 1 \le i \ne j \le n+1$ & $\checkmark$ \\ 
Lifelong learning & $\mathcal{S}^1, \cdots, \mathcal{S}^n$ & $\mathcal{S}^1, \cdots, \mathcal{S}^n$ &$\mathcal{S}^i$ arrives sequentially &$\checkmark$\\
Zero-shot learning & $\mathcal{S}^1, \cdots, \mathcal{S}^n$ & $\mathcal{S}^{n+1}$ & $\mathcal{Y}^{n+1} \ne \mathcal{Y}^i, 1 \le i \le n$ & $\times$ \\
Domain generalization & $\mathcal{S}^1, \cdots, \mathcal{S}^n$ & $\mathcal{S}^{n+1}$ & $P(\mathcal{S}^i) \ne P(\mathcal{S}^j), 1 \le i \ne j \le n+1$ & $\times$\\
\bottomrule
\end{tabular}
\vspace{-.1in}

\end{table*}

\textbf{Domain adaptation}~(DA)~\citep{wang2018deep,patel2015visual} is also popular in recent years.
DA aims to maximize the performance on a given target domain using existing training source domain(s).
The difference between DA and DG is that DA has access to the target domain data while DG cannot see them during training.
This makes DG more challenging than DA but more realistic and favorable in practical applications.

\textbf{Meta-learning}~\citep{vilalta2002perspective,hospedales2020meta,vanschoren2018meta} aims to learn the learning algorithm itself by learning from previous experience or tasks, i.e., learning-to-learn.
While the learning tasks are different in meta-learning, the learning tasks are the same in domain generalization.
Meta-learning is a general learning strategy that can be used for DG~\citep{li2018learning2,balaji2018metareg,li2019feature,Du2020LearningTL} by simulating the meta-train and meta-test tasks in training domains to enhance the performance of DG.

\textbf{Lifelong Learning}~\citep{biesialska2020continual}, or continual learning, cares about the learning ability among multiple sequential domains/tasks. It requires the model to continually learn over time by accommodating new knowledge
while retaining previously learned experiences.
This is also different from DG since it can access the target domain in each time step, and it does not explicitly handle the different distributions across domains.

\textbf{Zero-shot learning}~\citep{wang2019survey,ji2019decadal} aims at learning models from seen classes and classify samples whose categories are unseen in training. 
In contrast, domain generalization in general studies the problem where training and testing data are from the same classes but with different distributions.

\section{Theory}
\label{sec-theory}

\newcommand{\clA}{\mathcal{A}}
\newcommand{\clB}{\mathcal{B}}
\newcommand{\clE}{\mathcal{E}}
\newcommand{\clEh}{{\hat \clE}}
\newcommand{\clF}{\mathcal{F}}
\newcommand{\clH}{\mathcal{H}}
\newcommand{\clHt}{{\tilde \clH}}
\newcommand{\clP}{\mathcal{P}}
\newcommand{\clS}{\mathcal{S}}
\newcommand{\clU}{\mathcal{U}}
\newcommand{\clUt}{{\tilde \clU}}
\newcommand{\clX}{\mathcal{X}}
\newcommand{\clY}{\mathcal{Y}}
\newcommand{\clZ}{\mathcal{Z}}

\newcommand{\bbE}{\mathbb{E}}

\newcommand{\scX}{\mathscr{X}}

\newcommand{\frkk}{\mathfrak{k}}

\newcommand{\bfxx}{\mathbf{x}}

\newcommand{\ddh}{{\hat d}}
\newcommand{\epsilonh}{{\hat \epsilon}}

\newcommand{\kkb}{{\bar k}}

In this section, we review some theories related to \dg.
Since domain adaptation is closely related to DG, we begin with the theory of domain adaptation.

\subsection{Domain Adaptation}
\label{sec:thr-da}

For a binary classification problem, we denote the true labeling functions on the source domain as ${h^*}^s: \mathcal{X} \rightarrow [0,1]$
\footnote{When the output is in $(0,1)$, it means the probability of $y=1$.}
and that on the target domain as ${h^*}^t$.
Let $h: \mathcal{X} \rightarrow [0,1]$ be any classifier from a hypothesis space $\clH$.
The classification difference on the source domain between two classifiers $h$ and $h'$ can then be measured by
\begin{align}
    \epsilon^s (h,h') = \bbE_{\bfxx \sim P^s_X} [h(\bfxx) \neq h'(\bfxx)] = \bbE_{\bfxx \sim P^s_X} [|h(\bfxx) - h'(\bfxx)|],
\end{align}
and similarly we can define $\epsilon^t$ when taking $\bfxx \sim P^t_X$ in the expectation.
We define $\epsilon^s(h) := \epsilon^s(h, {h^*}^s)$ and $\epsilon^t(h) := \epsilon^t(h, {h^*}^t)$ as the risk of classifier $h$ on the source and target domains, respectively.

The goal of DG/DA is to minimize the target risk $\epsilon^t(h)$, but it is not accessible since we do not have any information on ${h^*}^t$.
So people seek to bound the target risk $\epsilon^t(h)$ using the tractable source risk $\epsilon^s(h)$.
Ben-David et al. \cite{ben2010theory} (Thm.~1) give a bound relating the two risks:
\begin{align}
  \epsilon^t(h) \le{} & \epsilon^s(h) + 2 d_1(P^s_X, P^t_X) \\
  & {}+ \min_{P_X \in \{P^s_X, P^t_X\}} \bbE_{\bfxx \sim P_X} [| {h^*}^s(\bfxx) - {h^*}^t(\bfxx) |],
  \label{eqn:da-bound-tv}
\end{align}
where $d_1(P^s_X, P^t_X) := \sup_{\clA \in \scX} |P^s_X[\clA] - P^t_X[\clA]|$ is the \emph{total variation} between the two distributions, and $\scX$ denotes the sigma-field on $\clX$.
The second term on the r.h.s measures the difference of cross-domain distributions,
and the third term represents the difference in the labeling functions
(covariate shift is not \emph{a priori} assumed).

However, the total variation is a strong distance (i.e., it tends to be very large) that may loosen the bound~\eqref{eqn:da-bound-tv}, and is hard to estimate using finite samples.
To address this, \citet{ben2010theory} developed another bound (\cite{ben2010theory}, Thm.~2; \cite{johansson2019support}, Thm.~1):
\begin{align}
  \epsilon^t(h) \le{} & \epsilon^s(h) + d_{\clH\Delta\clH}(P^s_X, P^t_X) + \lambda_\clH,
  \label{eqn:da-bound-hdh}
\end{align}
where the $\clH\Delta\clH$-divergence is defined as $d_{\clH\Delta\clH}(P^s_X, P^t_X) := \sup_{h, h' \in \clH} |\epsilon^s(h, h') - \epsilon^t(h, h')|$, replacing the total variation $d_1$ to measure the distribution difference,
and the ideal joint risk $\lambda_\clH := \inf_{h \in \clH} \left[ \epsilon^s(h) + \epsilon^t(h) \right]$ measures the complexity of $\clH$ for the prediction tasks on the two domains.
$\clH\Delta\clH$-divergence has a better finite-sample guarantee, leading to a non-asymptotic bound:
\begin{theorem}[Domain adaptation error bound (non-asymptotic) \cite{ben2010theory} (Thm.~2)]
\label{the-da-nonasymp}
Let $d$ be the Vapnik–Chervonenkis (VC) dimension~\cite{vapnik1994measuring} of $\clH$, and $\clU^s$ and $\clU^t$ be unlabeled samples of size $n$ from the two domains.
Then for any $h \in \clH$ and $\delta \in (0,1)$, the following inequality holds with probability at least $1-\delta$:
\begin{align}
    \epsilon^t(h) \le{} & \epsilon^s(h) + \ddh_{\clH\Delta\clH} (\clU^s, \clU^t) + \lambda_\clH \\
    & {}+ 4 \sqrt{\frac{ 2d \log (2n) + \log (2/\delta) }{n}},
\end{align}
where $\ddh_{\clH\Delta\clH}(\clU^s, \clU^t)$ is the estimate of $d_{\clH\Delta\clH}(P^s_X, P^t_X)$ on the two sets of finite data samples.
\end{theorem}

In the above bounds, the domain distribution difference $d(P^s_X, P^t_X)$ is not controllable, but one may learn a representation function $g: \clX \to \clZ$ that maps the original input data $\bfxx$ to some representation space $\clZ$, so that the representation distributions of the two domains become closer.
This direction of DA is thus called DA based on domain-invariant representation (DA-DIR).
The theory of domain-invariant representations has since inspired many DA/DG methods, which can be seen in Section~\ref{sec-repr}.

\subsection{Domain Generalization}

\subsubsection{Average risk estimation error bound}
The first line of \dg theory considers the case where the target domain is totally unknown (not even unsupervised data), and measures the average risk over all possible target domains.
Assume that all possible target distributions follow an underlying hyper-distribution $\clP$ on $(\bfxx,y)$ distributions: $P^t_{XY} \sim \clP$,
and that the source distributions also follow the same hyper-distribution: $P^1_{XY}, \cdots, P^M_{XY} \sim \clP$.
For generalization to any possible target domain, the classifier to be learned in this case also includes the domain information $P_X$ into its input, so prediction is in the form $y = h(P_X, \bfxx)$ on the domain with distribution $P_{XY}$.
For such a classifier $h$, its average risk over all possible target domains is then given by:
\begin{align}
    \clE(h) := \bbE_{P_{XY} \sim \clP} \bbE_{(\bfxx,y) \sim P_{XY}} [\ell(h(P_X, \bfxx), y)],
    \label{eqn:dg-avg-risk-exact}
\end{align}
where $\ell$ is a loss function on $\clY$.
Exactly evaluating the expectations is impossible, but we can estimate it using finite domains/distributions following $\clP$, and finite $(\bfxx,y)$ samples following each distribution.
As we have assumed $P^1_{XY}, \cdots, P^M_{XY} \sim \clP$, the source domains and supervised data could serve for this estimation:
\begin{align}
    \clEh(h) := \frac{1}{M} \sum_{i=1}^M \frac{1}{n^i} \sum_{j=1}^{n^i} \ell(h(\clU^i, \bfxx^i_j), y^i_j),
    \label{eqn:dg-avg-risk-est}
\end{align}
where we use the supervised dataset $\clU^i := \{\bfxx^i_j \mid (\bfxx^i_j, y^i_j) \in \clS^i\}$ from domain $i$ as an empirical estimation for $P^i_X$.

The first problem to consider is how well such an estimate approximates the target $\clE(h)$.
This can be measured by the largest difference between $\clE(h)$ and $\clEh(h)$ on some space of $h$.
To our knowledge, this is first analyzed by \citet{blanchard2011generalizing}, where the space of $h$ is taken as a reproducing kernel Hilbert space (RKHS).
However, different from common treatment, the classifier $h$ here also depends on the distribution $P_X$, so the kernel defining the RKHS should be in the form $\kkb((P_X^1, \bfxx_1), (P_X^2, \bfxx_2))$.
\citet{blanchard2011generalizing} construct such a kernel using kernels $k_X, k'_X$ on $\clX$ and kernel $\kappa$ on the RKHS $\clH_{k'_X}$ of kernel $k'_X$:
$\kkb((P_X^1, \bfxx_1), (P_X^2, \bfxx_2)) := \kappa(\Psi_{k'_X} (P_X^1), \Psi_{k'_X} (P_X^2)) k_X(\bfxx_1, \bfxx_2)$,
where $\Psi_{k'_X} (P_X) := \bbE_{\bfxx \sim P_X} [k'_X(\bfxx, \cdot)] \in \clH_{k'_X}$ is the kernel embedding of distribution $P_X$ via kernel $k'$.
The result is given in the following theorem, which gives a bound on the largest average risk estimation error within an origin-centered closed ball $\clB_{\clH_\kkb} (r)$ of radius $r$ in the RKHS $\clH_\kkb$ of kernel $\kkb$,
in a slightly simplified case where $n^1 = \cdots = n^M =: n$.
\begin{theorem}[Average risk estimation error bound for binary classification~\cite{blanchard2011generalizing}]
    Assume that the loss function $\ell$ is $L_\ell$-Lipschitz in its first argument and is bounded by $B_\ell$.
    Assume also that the kernels $k_X, k'_X$ and $\kappa$ are bounded by $B_k^2, B_{k'}^2 \ge 1$ and $B_\kappa^2$, respectively, and the canonical feature map $\Phi_\kappa: v \in \clH_{k'_X} \mapsto \kappa(v, \cdot) \in \clH_\kappa$ of $\kappa$ is $L_\kappa$-H\"older of order $\alpha \in (0,1]$ on the closed ball $\clB_{\clH_{k'_X}} (B_{k'})$ \footnote{
        This means that for any $u,v \in \clB_{\clH_{k'_X}} (\clB_{\clH_{k'}})$, it holds that $\Vert \Phi_\kappa(u) - \Phi_\kappa(v) \Vert \le L_\kappa \Vert u - v \Vert^\alpha$, where the norms are of the respective RKHSs.}.
    Then for any $r > 0$ and $\delta \in (0,1)$, with probability at least $1-\delta$, it holds that:
    \begin{align}
        \sup_{h \in \clB_{\clH_\kkb} (r)} \left| \clEh(h) - \clE(h) \right|
        \le{} & C \bigg( B_\ell \sqrt{- M^{-1} \log \delta} \\
        + r B_k L_\ell \Big( B_{k'} L_\kappa \big( & n^{-1} \log (M/\delta) \big)^{\alpha/2} + B_\kappa / \sqrt{M} \Big) \bigg),
    \end{align}
    where $C$ is a constant.
\end{theorem}
The bound becomes larger in general if $(M, n)$ is replaced with $(1, Mn)$.
It indicates that using domain-wise datasets is better than just pooling them into one mixed dataset, so the domain information plays a role.
This result is later extended in \cite{muandet2013domain}, and \citet{deshmukh2019generalization} give a bound for multi-class classification in a similar form.

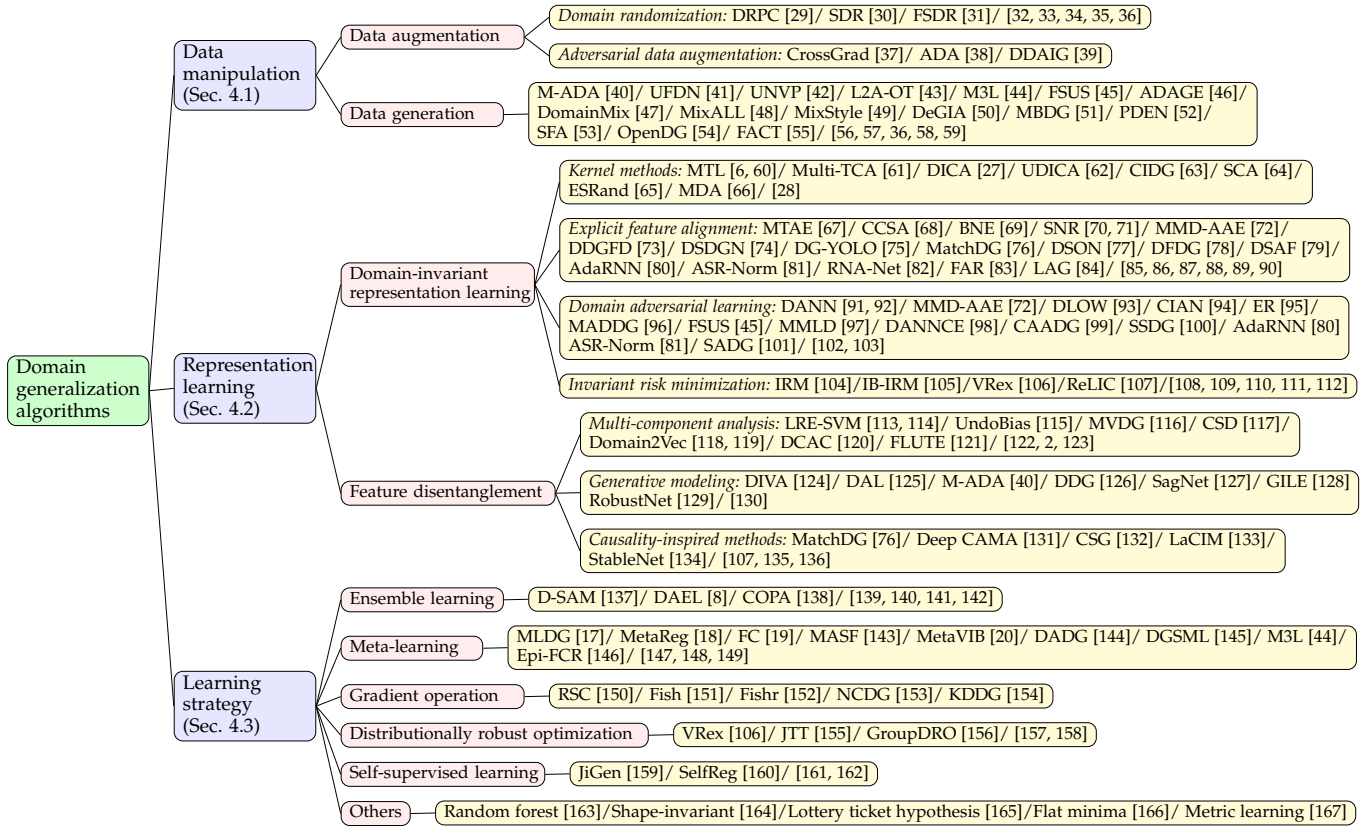
\begin{figure*}[t!]
	\centering
	\resizebox{\textwidth}{!}{
	\begin{forest}
  for tree={
  grow=east,
  reversed=true,
  anchor=base west,
  parent anchor=east,
  child anchor=west,
  base=left,
  font=\small,
  rectangle,
  draw,
  rounded corners,align=left,
  minimum width=2.5em,
  inner xsep=4pt,
  inner ysep=1pt,
  },
  where level=1{text width=5em,fill=blue!10}{},
  where level=2{text width=5em,font=\footnotesize,fill=pink!30}{},
  where level=3{font=\footnotesize,yshift=0.26pt,fill=yellow!20}{},
  [Domain \\ generalization\\algorithms,fill=green!20
        [Data\\manipulation\\(Sec.~\ref{sec-method-data}),text width=6em
            [Data augmentation,text width=8em
              [\emph{Domain randomization:} DRPC \cite{yue2019domain}/ SDR \cite{prakash2019structured}/ FSDR \cite{huang2021fsdr}/ \cite{tobin2017domain,peng2018sim,khirodkar2019domain,tremblay2018training,nazari2020domain}
              ]
              [\emph{Adversarial data augmentation:} CrossGrad \cite{shankar2018generalizing}/ ADA \cite{volpi2018generalizing}/ DDAIG \cite{zhou2020deep}
              ]
            ]
            [Data generation,text width=7em [M-ADA \cite{qiao2020learning}/ UFDN \cite{liu2018unified}/ UNVP \cite{truong2019recognition}/ L2A-OT \cite{Zhou2020LearningTG}/ M3L \cite{zhao2020learning}/  FSUS \cite{garg2020learn}/ ADAGE \cite{maria2019hallucinating}/ \\ DomainMix \cite{wang2020domainmix}/ MixALL \cite{wang2020heterogeneous}/ MixStyle \cite{zhoudomain}/ DeGIA \cite{truong2019image}/  MBDG \cite{robey2021model}/ PDEN \cite{li2021progressive}/ \\ SFA \cite{li2021simple}/ OpenDG \cite{shu2021open}/  FACT \cite{xu2021fourier}/ \cite{rahman2019multi,somavarapu2020frustratingly,nazari2020domain,peng2021out,wang2021learningtodiversify}
              ]
            ]
        ]
        [Representation\\learning\\(Sec.~\ref{sec-repr}),text width=6em
          [Domain-invariant\\representation learning,text width=8.5em
              [\emph{Kernel methods:} MTL \cite{blanchard2011generalizing,blanchard2017domain}/ Multi-TCA \cite{grubinger2015domain}/ DICA \cite{muandet2013domain}/ UDICA \cite{gan2016learning}/ CIDG \cite{li2018domain}/ SCA \cite{ghifary2016scatter}/ \\ ESRand \cite{erfani2016robust}/ MDA \cite{hu2019domain}/ \cite{deshmukh2019generalization}
              ]
              [\emph{Explicit feature alignment:} MTAE \cite{Ghifary2015DomainGF}/ CCSA \cite{motiian2017unified}/ BNE \cite{segu2020batch}/ SNR \cite{jin2020style,jin2021style}/ MMD-AAE \cite{Li2018DomainGW}/ \\ DDGFD \cite{zheng2020deep}/ DSDGN \cite{liao2020deep}/  DG-YOLO \cite{liu2020towards}/ MatchDG \cite{Mahajan2020DomainGU}/  DSON \cite{seo2020learning}/  DFDG \cite{zhang2021robust}/ DSAF \cite{qi2021unsupervised}/ \\ AdaRNN \cite{du2021adarnn}/ ASR-Norm \cite{fan2021adversarially}/ RNA-Net \cite{planamente2021domain}/ FAR \cite{jin2020feature}/ LAG \cite{lu2022local}/ \cite{maniyar2020zero,han2020domain,gulrajani2021search,ayodele2020supervised,chen2020study,li2020domain}
              ]
              [\emph{Domain adversarial learning:} DANN \cite{ganin2015unsupervised,Ganin2016DomainAdversarialTO}/ MMD-AAE \cite{Li2018DomainGW}/ DLOW \cite{gong2019dlow}/ CIAN \cite{li2018deep}/ ER \cite{zhao2020domain}/ \\ MADDG \cite{shao2019multi}/  FSUS \cite{garg2020learn}/ MMLD \cite{Matsuura2020DomainGU}/ DANNCE \cite{sicilia2021domain}/ CAADG \cite{rahman2020correlation}/ SSDG \cite{jia2020single}/ AdaRNN \cite{du2021adarnn} \\ ASR-Norm \cite{fan2021adversarially}/ SADG \cite{luo2021scale}/ \cite{albuquerque2019adversarial,wang2020unseen}
              ]
              [\emph{Invariant risk minimization:} IRM \cite{arjovsky2019invariant}/IB-IRM \cite{ahuja2021invariance}/VRex \cite{krueger2021out}/ReLIC \cite{mitrovic2021representation}/\cite{guo2021out,ahuja2020empirical,choe2020empirical,rosenfeld2020risks,pmlr-v119-ahuja20a}
              ]
          ]
          [Feature disentanglement, text width=9.5em
            [\emph{Multi-component analysis:} LRE-SVM \cite{xu2014exploiting,li2017domain}/ UndoBias \cite{khosla2012undoing}/ MVDG \cite{niu2015multi}/ CSD \cite{piratla2020efficient}/ \\ Domain2Vec \cite{deshmukh2018domain2vec,peng2020domain2vec}/   DCAC \cite{hu2021domain}/ FLUTE \cite{triantafillou2021learning}/  \cite{ding2017deep,li2017deeper,zunino2020explainable}
            ]
            [\emph{Generative modeling:} DIVA \cite{ilse2019diva}/ DAL \cite{peng2019domain}/ M-ADA \cite{qiao2020learning}/ DDG \cite{zhang2021towards}/ SagNet \cite{nam2021reducing}/ GILE \cite{qian2021latent} \\ RobustNet \cite{choi2021robustnet}/ \cite{li2021semantic}
            ]
            [\emph{Causality-inspired methods:} MatchDG \cite{Mahajan2020DomainGU}/ Deep CAMA \cite{zhang2020causal}/ CSG \cite{liu2021learning}/  LaCIM \cite{sun2021recovering}/ \\ StableNet \cite{zhang2021deep}/ \cite{mitrovic2021representation,ouyang2021causality,he2019towards}
            ]
          ]
        ]
        [Learning\\strategy\\(Sec.~\ref{sec-method-learn}),text width=6em
          [Ensemble learning,text width=7em
            [D-SAM \cite{d2018domain}/ DAEL \cite{zhou2020domain}/ COPA \cite{wu2021collaborative}/ \cite{mancini2018best,he2018domain,mancini2018robust,dubey2021adaptive}]
            ]
          [Meta-learning, text width=6em
            [MLDG \cite{li2018learning2}/ MetaReg \cite{balaji2018metareg}/ FC \cite{li2019feature}/ MASF \cite{Dou2019DomainGV}/ MetaVIB \cite{Du2020LearningTL}/ DADG \cite{chen2020discriminative}/ DGSML \cite{sharifi2020domain}/ M3L \cite{zhao2020learning}/ \\ Epi-FCR \cite{li2019episodic}/ \cite{wang2020meta,qiao2021uncertainty,kim2021self}]
            ]
         [Gradient operation, text width=8em
            [RSC \cite{huang2020self}/ Fish \cite{shi2021gradient}/ Fishr \cite{rame2021fishr}/ NCDG \cite{tian2021neuron}/ KDDG \cite{wang2021embracing}]
            ]
         [Distributionally robust optimization, text width=14em
            [VRex \cite{krueger2021out}/ JTT \cite{liu2021just}/ GroupDRO \cite{sagawa2019distributionally}/ \cite{koh2020wilds,wang2021class}]
            ]
         [Self-supervised learning, text width=9em
            [JiGen \cite{carlucci2019domain}/ SelfReg \cite{kim2021selfreg}/ \cite{jeon2021feature,li2021domain}]
            ]
          [Others, text width=2.5em
                [Random forest~\cite{ryu2019generalized}/Shape-invariant~\cite{narayanan2021shape}/Lottery ticket hypothesis~\cite{zhang2021can}/Flat minima~\cite{cha2021swad}/
                Metric learning~\cite{faraki2021cross}
                ]
            ]
        ]
    ]
\end{forest}
	}
	\caption{Taxonomy of domain generalization methods.}
	\label{fig-main}
\end{figure*}

\subsubsection{Generalization risk bound}
Another line of DG theory considers the risk on a specific target domain, under the assumption of covariate shift (i.e., the labeling function $h^*$ or $P_{Y|X}$ is the same over all domains).
This measurement is similar to what is considered in domain adaptation theory in Section~\ref{sec:thr-da}, so we adopt the same definition for the source risks $\epsilon^1, \cdots, \epsilon^M$ and the target risk $\epsilon^t$.
With the covariate shift assumption, each domain is characterized by the distribution on $\clX$.
\citet{albuquerque2019adversarial} then consider approximating the target domain distribution $P^t_X$ within the convex hull of source domain distributions: $\Lambda := \{\sum_{i=1}^M \pi_i P^i_X \mid \pi \in \Delta_M\}$, where $\Delta_M$ is the $(M-1)$-dimensional simplex so that each $\pi$ represents a normalized mixing weights.
Similar to the domain adaptation case, distribution difference is measured by the $\clH$-divergence to include the influence of the classifier class.
\begin{theorem}[Domain generalization error bound~\cite{albuquerque2019adversarial}]
    Let $\gamma := \min_{\pi \in \Delta_M} d_\clH (P^t_X, \sum_{i=1}^M \pi_i P^i_X)$ with minimizer $\pi^*$
    \footnote{The original presentation does not mention that the $\pi$ is the minimizer, but the proof indicates so.}
    be the distance of $P^t_X$ from the convex hull $\Lambda$, and $P^*_X := \sum_{i=1}^M \pi^*_i P^i_X$ be the best approximator within $\Lambda$.
    Let $\rho := \sup_{P'_X, P''_X \in \Lambda} d_\clH (P'_X, P''_X)$ be the diameter of $\Lambda$.
    Then it holds that
    \begin{align}
        \epsilon^t(h) \le \sum_{i=1}^M \pi^*_i \epsilon^i(h) + \frac{\gamma + \rho}{2} + \lambda_{\clH, (P^t_X, P^*_X)},
    \end{align}
    where $\lambda_{\clH, (P^t_X, P^*_X)}$ is the ideal joint risk across the target domain and the domain with the best approximator distribution $P^*_X$.
\end{theorem}
The result can be seen as the generalization of domain adaptation bounds in Section~\ref{sec:thr-da} when there are multiple source domains.
Again similar to the domain adaptation case, this bound motivates \dg methods based on domain invariant representation, which simultaneously minimize the risks over all source domains corresponding to the first term of the bound, as well as the representation distribution differences among source and target domains in the hope to reduce $\gamma$ and $\rho$ on the representation space.
To sum up, the theory of generalization is an active research area and other researchers also derived different DG theory bounds using informativeness~\cite{ye2021towards} and adversarial training~\cite{albuquerque2019adversarial,sicilia2021domain,ye2021towards,deshmukh2019generalization}.

\section{Methodology}
\label{sec-method}

In this section, we introduce existing domain generalization methods in detail.
As shown in \figurename~\ref{fig-main}, we categorize them into three groups, namely:

\begin{enumerate}[(1)]
    \item \textbf{Data manipulation:} This category of methods focuses on manipulating the inputs to assist learning general representations. Along this line, there are two kinds of popular techniques: a). \emph{Data augmentation}, which is mainly based on augmentation, randomization, and transformation of input data; b). \emph{Data generation}, which generates diverse samples to help generalization. 
    
    \item \textbf{Representation learning:} This category of methods is the most popular in \dg. There are two representative techniques: a). \emph{Domain-invariant representation learning},
    which performs kernel, adversarial training, explicitly feature alignment between domains, or invariant risk minimization to learn domain-invariant representations; b). \emph{Feature disentanglement}, which tries to disentangle the features into domain-shared or domain-specific parts for better generalization.
    
    \item \textbf{Learning strategy:} This category of methods focuses on exploiting the general learning strategy to promote the generalization capability, which mainly includes several kinds of methods: a). \emph{Ensemble learning}, which relies on the power of ensemble to learn a unified and generalized predictive function; b). \emph{Meta-learning}, which is based on the learning-to-learn mechanism to learn general knowledge by constructing meta-learning tasks to simulate domain shift; c). \emph{Gradient operation}, which tries to learn generalized representations by directly operating on gradients; d). \emph{Distributionally robust optimization}, which learns the worst-case distribution scenario of training domains; e). \emph{Self-supervised learning}, which constructs pretext tasks to learn generalized representations. Additionally, there are other learning strategy that can be used for DG.
\end{enumerate}

These three categories of approaches are conceptually different. They are complementary to each other and can be combined towards higher performance.
We will describe the approaches for each category in detail hereafter.

\subsection{Data Manipulation}
\label{sec-method-data}
We are always hungry for more training data in machine learning (ML).
The generalization performance of a ML model often relies on the quantity and diversity of the training data.
Given a limited set of training data, data manipulation is one of the cheapest and simplest way to generate samples so as to enhance the generalization capability of the model.
The main objective for data manipulation-based DG is to increase the diversity of existing training data using different data manipulation methods. At the same time, the data quantity is also increased.
Although the theoretical insight for why data augmentation or generation techniques can enhance the generalization ability of a model, experiments by \citet{adila2021understanding} showed that the model tend to make predictions for both OOD and in-distribution samples based on trivial syntactic heuristics for NLP tasks.

We formulate the general learning objective of data manipulation-based DG as:
\begin{equation}
    \label{eq-augmentation-all}
    \min_{h} \, \mathbb{E}_{\mathbf{x},y} [ \ell(h(\mathbf{x}),y) ] + \mathbb{E}_{\mathbf{x}^\prime,y} [ \ell(h(\mathbf{x}^\prime),y) ],
\end{equation}
where $\mathbf{x}^\prime = \mathcal{M}(\mathbf{x})$ denotes the manipulated data using a function $\mathcal{M}(\cdot)$.
Based on the difference on this function, we further categorize existing work into two types: \emph{data augmentation} and \emph{data generation}.

\subsubsection{Data augmentation-based DG}
Augmentation is one of the most useful techniques for training machine learning models.
Typical augmentation operations include flipping, rotation, scaling, cropping, adding noise, and so on.
They have been widely used in supervised learning to enhance the generalization performance of a model by reducing overfitting \cite{shorten2019survey,nazari2020domain}.
Without exception, they can also be adopted for DG where $\mathcal{M}(\cdot)$ can be instantiated as these data augmentation functions.

\paragraph{\textbf{Domain randomization}}
Other than typical augmentation, domain randomization is an effective technique for data augmentation.
It is commonly done by generating new data that can simulate complex environments based on the limited training samples.
Here, the $\mathcal{M}(\cdot)$ function is implemented as several manual transformations (commonly used in image data) such as: altering the location and texture of objects, changing the number and shape of objects, modifying the illumination and camera view, and adding different types of random noise to the data.
\citet{tobin2017domain} first used this method to generate more training data from the simulated environment for generalization in the real environment.
Similar techniques were also used in \cite{peng2018sim,khirodkar2019domain,tremblay2018training,yue2019domain} to strengthen the generalization capability of the models.
\citet{prakash2019structured} further took into account the structure
of the scene when randomly placing objects for data generation, which enables the neural network to learn to utilize context when detecting objects.
\citet{peng2021out} proposed to not only augment features, but also labels.
It is easy to see that by randomization, the diversity of samples can be increased.
But randomization is often random, indicating that there could be some useless randomizations that could be further removed to improve the efficiency of the model.
%


\paragraph{\textbf{Adversarial data augmentation}}
Adversarial data augmentation aims to guide the augmentation to optimize the generalization capability, by enhancing the diversity of data while assuring their reliability. 
\citet{shankar2018generalizing} used a Bayesian network to model dependence between label, domain and input instance, and proposed CrossGrad, a cautious data augmentation strategy that perturbs the input along the direction of greatest domain change while changing the class label as little as possible. 
\citet{volpi2018generalizing} proposed an iterative procedure that augments the source dataset with examples from a fictitious target domain that is ``hard" under the current model, where adversarial examples are appended at each iteration to enable adaptive data augmentation. 
\citet{zhou2020deep} adversarially trained a transformation network for data augmentation instead of directly updating the inputs by gradient ascent while they adopted the regularization of weak and strong augmentation in \cite{zhou2021semi,huang2021fsdr}.
Adversarial data augmentation often has certain optimization goals that can be used by the network.
However, its optimization process often involves adversarial training, thus is difficult.

\subsubsection{Data generation-based DG}
Data generation is also a popular technique to generate diverse and rich data to boost the generalization capability of a model.
Here, the function $\mathcal{M}(\cdot)$ can be implemented using some generative models such as Variational Auto-encoder (VAE)~\cite{kingma2013auto}, and Generative Adversarial Networks (GAN)~\cite{goodfellow2014generative}.
In addition, it can also be implemented using the Mixup~\cite{zhang2018mixup} strategy.

\citet{rahman2019multi} used ComboGAN~\cite{anoosheh2018combogan} to generate new data and then applied domain discrepancy measure such as MMD~\cite{gretton2012kernel} to minimize the distribution divergence between real and generated images to help learn general representations.
\citet{qiao2020learning} leveraged adversarial training to create ``fictitious" yet ``challenging" populations, where a Wasserstein Auto-Encoder (WAE)~\cite{tolstikhin2017wasserstein} was used to help generate samples that preserve the semantic and have large domain transportation.
\citet{Zhou2020LearningTG} generated novel distributions under semantic consistency and then maximized the difference between source and the novel distributions.
\citet{somavarapu2020frustratingly} introduced a simple transformation based on image stylization to explore cross-source variability for better generalization, where AdaIN \cite{huang2017arbitrary} was employed to achieve fast stylization to arbitrary styles.
Different from others, \citet{li2021progressive} used adversarial training to generate \emph{domains} instead of samples.
These methods are more complex since different generative models are involved and we should pay attention to the model capacity and computing cost.


In addition to the above generative models, Mixup~\citep{zhang2018mixup} is also a popular technique for data generation.
Mixup generates new data by performing linear interpolation between any two instances and between their labels with a weight sampled from a Beta distribution, which does not require to train generative models.
Recently, there are several methods using Mixup for DG, by either performing Mixup in the original space~\citep{wang2020domainmix,wang2020heterogeneous,shu2021open} to generate new samples; or in the feature space~\citep{zhoudomain,qiao2021uncertainty,xu2021fourier} which does not explicitly generate raw training samples.
These methods achieved promising performance on popular benchmarks while remaining conceptually and computationally simple.

\subsection{Representation Learning}
\label{sec-repr}


Representation learning has always been the focus of machine learning for decades~\cite{bengio2013representation} and is also one of the keys to the success of \dg. 
We decompose the prediction function $h$ as $h = f \circ g$, where $g$ is a representation learning function and $f$ is the classifier function. The goal of representation learning can be formulated as:
\begin{equation}
    \label{eq-repre-all}
    \min_{f,g} \, \mathbb{E}_{\mathbf{x},y}  \ell(f(g(\mathbf{x})),y)  + \lambda \ell_{\operatorname{reg}},
\end{equation}
where $\ell_{\operatorname{reg}}$ denotes some regularization term and $\lambda$ is the tradeoff parameter.
Many methods are designed to better learn the feature extraction function $g$ with corresponding $\ell_{\operatorname{reg}}$. In this section, we categorize the existing literature on representation learning into two main categories based on different learning principles: \emph{domain-invariant representation learning} and \emph{feature disentanglement}.

\subsubsection{Domain-invariant representation-based DG}

The work of \cite{ben2007analysis} theoretically proved that if the feature representations remain invariant to different domains, the representations are general and transferable to different domains (also refer to Section~\ref{sec-theory}).
Based on this theory, a plethora of algorithms have been proposed for \da.
Similarly, for \dg, the goal is to reduce the representation discrepancy between multiple source domains in a specific feature space to be domain invariant so that the learned model can have a generalizable capability to the unseen domain. 
Along this line, there are mainly four types of methods: kernel-based methods, domain adversarial learning, explicit feature alignment, and invariant risk minimization.

\paragraph{\textbf{Kernel-based methods}}
Kernel-based method is one of the most classical learning paradigms in machine learning.
Kernel-based machine learning relies on the kernel function to transform the original data into a high-dimensional feature space without ever computing the coordinates of the data in that space, but by simply computing the inner products between the samples of all pairs in the feature space.
One of the most representative kernel-based methods is Support Vector Machine (SVM)~\cite{cortes1995support}.
For \dg, there are plenty of algorithms based on kernel methods, where the representation learning function $g$ is implemented as some feature map $\phi(\cdot)$ which is easily computed using kernel function $k(\cdot, \cdot)$ such as RBF kernel and Laplacian kernel.

\citet{blanchard2011generalizing} first used kernel method for \dg  and extended it in \cite{blanchard2017domain}. They adopted the positive semi-definite kernel learning to learn a domain-invariant kernel from the training data.
\citet{grubinger2015domain} adapted transfer component analysis (TCA)~\cite{Pan2011DomainAV} to bridge the multi-domain distance to be closer for DG.
Similar to the core idea of TCA, Domain-Invariant Component Analysis (DICA)~\cite{muandet2013domain} is one of the classic methods using kernel for DG.
The goal of DICA is to find a feature transformation kernel $k(\cdot, \cdot)$ that minimizes the distribution discrepancy between all data in feature space.
\citet{gan2016learning} adopted a similar method as DICA and further added attribute regularization.
In contrast to DICA which deals with the marginal distribution, \citet{li2018domain} learned a feature representation which has domain-invariant class conditional distribution.
Scatter component analysis (SCA)~\cite{ghifary2016scatter} adopted Fisher's discriminant analysis to minimize the discrepancy of representations from the same class and the same domain, and maximize the discrepancy of representations from the different classes and different domains.
\citet{erfani2016robust} proposed an Elliptical Summary Randomisation (ESRand) that comprises of a randomised kernel and elliptical data summarization.
ESRand projected each domain into an ellipse to represent the domain information and then used some similarity metric to compute the distance.
\citet{hu2019domain} proposed multi-domain discriminant analysis to perform class-wise kernel learning for DG, which is more fine-grained.
To sum up, this category of methods is often highly related to other categories to act as their divergence measures or theoretical support.


\paragraph{\textbf{Domain adversarial learning}}
Domain-adversarial training is widely used for learning domain-invariant features.
\citet{ganin2015unsupervised} and \citet{Ganin2016DomainAdversarialTO} proposed Domain-adversarial neural network (DANN) for domain adaptation, which adversarially trains the generator and discriminator. The discriminator is trained to distinguish the domains while the generator is trained to fool the discriminator to learn domain invariant feature representations. 
\citet{Li2018DomainGW} adopted such idea for DG.
\citet{gong2019dlow} used adversarial training by gradually reducing the domain discrepancy in a manifold space.
\citet{li2018deep} proposed a conditional invariant adversarial network (CIAN) to learn class-wise adversarial networks for DG.
Similar ideas were also used in \cite{shao2019multi,rahman2020correlation,wang2020unseen}.
\citet{jia2020single} used single-side adversarial learning and asymmetric triplet loss to make sure only the real faces from different domains were indistinguishable, but not for the fake ones. After that, the extracted features of fake faces are more dispersed than before in the feature space and those of real ones are more aggregated, leading to a better generalized class boundary for unseen domains.
In addition to adversarial domain classification, \citet{zhao2020domain} introduced additional entropy regularization by minimizing the KL divergence between the conditional distributions of different training domains to push the network to learn domain-invariant features. 
Some other GAN-based methods ~\cite{garg2020learn,sicilia2021domain,albuquerque2019adversarial} were also proposed with theoretically guaranteed generalization bound.


\paragraph{\textbf{Explicit feature alignment}}
This line of works aligns the features across source domains to learn domain-invariant representations through explicit feature distribution alignment \cite{Li2018DomainGW,Zhou2020DomainGW,peng2019moment,zhu2021self}, or feature normalization \cite{pan2018two,jia2019frustratingly,nam2018batch,jin2020style}.
\citet{motiian2017unified} introduced a cross-domain contrastive loss for representation learning, where mapped domains are semantically aligned and yet maximally separated.
Some methods explicitly minimized the feature distribution divergence by minimizing the maximum mean discrepancy (MMD) \cite{tzeng2014deep,Pan2011DomainAV,wang2018visual,wang2020transfer}, second order correlation \cite{sun2016return,sun2016deep,peng2018synthetic}, both mean and variance (moment matching) \cite{peng2019moment}, Wasserstein distance \cite{Zhou2020DomainGW}, \etcno, of domains for either \da or \dg.
\citet{Zhou2020DomainGW} aligned the marginal distribution of different source domains via optimal transport by minimizing the Wasserstein distance to achieve domain-invariant feature space.

Moreover, there are some works that used feature normalization techniques to enhance domain generalization capability \cite{pan2018two,jia2019frustratingly}.
\citet{pan2018two} introduced Instance Normalization (IN) layers to CNNs to improve the generalization capability of models. IN has been extensively investigated in the field of image style transfer \cite{ulyanov2017improved,dumoulin2016learned,huang2017arbitrary}, where the style of an image is reflected by the IN parameters, \ieno, mean and variance of each feature channel. Thus, IN layers \cite{ulyanov2016instance} could be used to eliminate instance-specific style discrepancy to enhance generalization \cite{pan2018two}. However, IN is task agnostic and may remove some discriminative information. In IBNNet, IN and Batch Normalization (BN) are utilized in parallel to preserve some discriminative information \cite{pan2018two}. In \cite{nam2018batch}, BN layers are replaced by Batch-Instance Normalization (BIN) layers, which adaptively balance BN and IN for each channel by selectively using BN and IN.
\citet{jin2020style,jin2021style} proposed a Style Normalization and Restitution (SNR) module to simultaneously ensure both high generalization and discrimination capability of the networks. After the style normalization by IN, a restitution step is performed to distill task-relevant discriminative features from the residual (\ieno, the difference between the original feature and the style normalized feature) and add them back to the network to ensure high discrimination. The idea of restitution is extended to other alignment-based method to restorate helpful discriminative information dropped by alignment \cite{jin2020feature}.
Recently, \citet{qi2021unsupervised} applied IN to unsupervised DG where there are no labels in the training domains to acquire invariant and transferable features.
A combination of different normalization techniques is presented in \cite{fan2021adversarially} to show that adaptively learning the normalization technique can improve DG.
This category of methods is more flexible and can be applied to other kind of categories.

\paragraph{\textbf{Invariant risk minimization (IRM)}}
\newcommand{\clG}{\mathcal{G}}
\citet{arjovsky2019invariant} considered another perspective on the domain-invariance of representation for domain generalization.
They did not seek to match the representation distribution of all domains, but to enforce the optimal classifier on top of the representation space to be the same across all domains.
The intuition is that the ideal representation for prediction is the \emph{cause} of $y$, and the causal mechanism should not be affected by other factors/mechanisms, thus is domain-invariant.
Formally, IRM can be formulated as:
\begin{align}
    \min_{\substack{g \in \clG, \\
        f \in \bigcap_{i=1}^M \arg\min_{f' \in \clF} \epsilon^i (f' \circ g)}}
    \sum_{i=1}^M \epsilon^i (f \circ g)
\end{align}
for some function classes $\clF$ of $g$ and $\clF$ of $f$.
The constraint for $f$ embodies the desideratum that all domains share the same representation-level classifier, and the objective function encourages $f$ and $g$ to achieve a low source domain risk.
However, this problem is hard to solve as it involves an inner-level optimization problem in its constraint.
The authors then develop a surrogate problem to learn the feature extractor $g$ that is much more practical:
\begin{align}
    \min_{g \in \clG} \sum_{i=1}^M \epsilon^i (g) + \lambda \left\| \left. \nabla_f \epsilon^i (f \circ g) \right|_{f=1} \right\|^2,
\end{align}
where a dummy representation-level classifier $f = 1$ is considered, and the gradient norm term measures the optimality of this classifier.
The work also presents a generalization theory under a perhaps strong linear assumption, that for plenty enough source domains, the ground-truth invariant classifier can be identified.

IRM has gain notable visibility recently.
There are some further theoretical analyses on the success~\cite{pmlr-v119-ahuja20a} and failure cases of IRM~\cite{rosenfeld2020risks},
and IRM has been extended to other tasks including text classification~\cite{choe2020empirical} and reinforcement learning~\cite{sonar2020invariant}.
The idea to pursue the invariance of optimal representation-level classifier is also extended.
\citet{krueger2021out} promote this invariance by minimizing the extrapolated risk among source domains, which essentially minimizes the variance of source-domain risks.
\citet{mitrovic2021representation} aim to learn such a representation in a self-supervised setup, where the second domain is constructed by data augmentation showing various semantic-irrelevant variations.
Recently, \citet{ahuja2021invariance} found the invariance of $f$ alone is not sufficient. They found IRM still fails if $g$ captures ``fully informative invariant features'', which makes $y$ independent of $x$ on all domains. This is particularly the case for classification (vs. regression) tasks. An information bottleneck regularization is hence introduced to maintain only partially informative features.

\subsubsection{Feature disentanglement-based DG}

Disentangled representation learning aims to learn a function that maps a sample to a feature vector, which contains all the information about different factors of variation and each dimension (or a subset of dimensions) contains information about only some factor(s).
Disentanglement based DG approaches in general decompose a feature representation into understandable compositions/sub-features, with one feature being domain-shared/invariant feature and the other domain-specific feature.
The optimization objective of disentanglement-based DG can be summarized as:
\begin{equation}
    \label{eq-disent}
    \min_{g_c, g_s, f} \mathbb{E}_{\mathbf{x},y} \ell (f(g_c(\mathbf{x})), y) + \lambda \ell_{\mathrm{reg}} + \mu \ell_{\mathrm{recon}} ([ g_c(\mathbf{x}), g_s(\mathbf{x}) ], \mathbf{x} ),
\end{equation}
where $g_c$ and $g_s$ denote the domain-shared and domain-specific feature representations, respectively.
$\lambda$ and $\mu$ are tradeoff parameters.
The loss $\ell_{\mathrm{reg}}$ is a regularization term that explicitly encourages the separation of the domain shared and specific features and $\ell_{\mathrm{recon}}$ denotes a reconstruction loss that prevents information loss.
Note that $[ g_c(\mathbf{x}), g_s(\mathbf{x}) ]$ denotes the combination of two kinds of features (which is not limited to concatenation operation).

Based on the choice of network structures and implementation mechanisms, disentanglement-based DG can mainly be categorized into three types: \emph{multi-component analysis}, \emph{generative modeling}, and \emph{causality-inspired methods}.

\paragraph{\textbf{Multi-component analysis}}
In multi-component analysis, the domain-shared and domain-specific features are in general extracted using the domain-shared and domain-specific network parameters.
The method of UndoBias~\cite{khosla2012undoing} started from a SVM model to maximize interval classification on all training data for \dg.
They represented the parameters of the $i$-th domain as $\mathbf{w}_i=\mathbf{w}_0+\Delta_i$, where $\mathbf{w}_0$ denotes the domain-shared parameters and $\Delta_i$ denotes the domain-specific parameters.
Some other methods extented the idea of UndoBias from different aspects.
\citet{niu2015multi} proposed to use multi-view learning for \dg. They proposed Multi-view DG (MVDG) to learn the combination of exemplar SVMs under different views for robust generalization.
\citet{ding2017deep} designed domain-specific networks for each domain and one shared domain-invariant network for all domains to learn disentangled representations, where low-rank reconstruction is adopted to align two types of networks in structured low-rank fashion.
\citet{li2017deeper} extended the idea of UndoBias into the neural network context and developed a low-rank parameterized CNN model for end-to-end training.
\citet{zunino2020explainable} learned disentangled representations through manually comparing the attention heat maps for certain areas from different domains.
There are also other works that adopt multi-component analysis for disentanglement~\cite{deshmukh2018domain2vec,xu2014exploiting,li2017domain,piratla2020efficient,triantafillou2021learning, bui2021exploiting,kang2021dynamically,liu2021domain,peng2020domain2vec}.
In general, multi-component analysis can be implemented in different architectures and remains effective for representation disentanglement.

\paragraph{\textbf{Generative modeling}}
Generative models can be used for disentanglement from the perspective of data generation process.
This kind of methods tries to formulate the generative mechanism of the samples from the domain-level, sample-level, and label-level.
Some works further disentangle the input into class-irrelevant features, which contain the information related to specific instance~\cite{wang2021variational}.
The Domain-invariant variational autoencoder (DIVA)~\cite{ilse2019diva} disentangled the features into domain information, category information, and other information, which is learned in the VAE framework.
\citet{peng2019domain} disentangled the fine-grained domain information and category information that are learned in VAEs.
\citet{qiao2020learning} also used VAE for disentanglement, where they proposed a Unified Feature Disentanglement Network (UFDN) that treated both data domains and image attributes of interest as latent factors to be disentangled.
Similarly, \citet{zhang2021towards} disentangled the semantic and variational part of the samples.
Similar spirits also include \cite{li2021semantic,choi2021robustnet}.
\citet{nam2021reducing} proposed to disentangle the style and other information using generative models that their method is both for \da and \dg.
Generative models can not only improve OOD performance, but can also be used for generation tasks, which we believe is useful for many potential applications.

\paragraph{\textbf{Causality-inspired methods}}

Causality is a finer description of variable relations beyond statistics (joint distribution). Causal relation gives information of how the system will behave under intervention, so it is naturally suitable for transfer learning tasks, since domain shift can be seen as an intervention. Particularly, under the causal consideration, the desired representation is the true cause of the label (e.g., object shape), so that prediction will not be affected by intervention on correlated but semantically irrelevant features (e.g., background, color, style). There are a number of works~\citep{scholkopf2012causal,zhang2013domain,zhang2015multi,gong2018causal} that exploited causality for domain adaptation.

For domain generalization, \citet{he2019towards} reweighted input samples in a way to make the weighted correlation reflect causal effect. \citet{zhang2021deep} took Fourier features as the causing factors of images, and enforce the independence among these features. Using the additional data of object identity (it is a more detailed label than the class label), \citep{heinze2019conditional} enforced the conditional independence of the representation from domain index given the same object. When such object label is unavailable, \citep{Mahajan2020DomainGU} further learned an object feature based on labels in a separate stage. For single-source domain generalization, \citep{mitrovic2021representation,ouyang2021causality} used data augmentation to present information of the causal factor. The augmentation operation is seen as producing outcomes under intervention on irrelevant features, which is implemented based on specific domain knowledge.
There are also generative methods under the causal consideration. \citet{zhang2020causal} explicitly modeled a manipulation variable that causes domain shift, which may be unobserved.
\citet{liu2021learning} leveraged causal invariance for single-source generalization, i.e., the invariance of the process of generating $(x,y)$ data based on the factors, which is explained more general than inference invariance that existing methods implicitly rely on. The two factors are allowed correlated which is more realistic. They theoretically prove the identifiability of the causal factor is possible and the identification benefits generalization. \citet{sun2021recovering} extended the method and theory to multiple source domains. With more informative data, the irrelevant factor is also identifiable.


\subsection{Learning Strategy}
\label{sec-method-learn}
In addition to data manipulation and representation learning, DG was also studied in general machine learning paradigms, which is divided into several categories: \emph{ensemble learning-based DG}, \emph{meta-learning-based DG}, \emph{gradient operation-based DG}, \emph{distributionally robust optimization-based DG}, \emph{self-supervised learning-based DG}, and \emph{other strategies}.

\subsubsection{Ensemble learning-based DG}
Ensemble learning usually combines multiple models, such as classifiers or experts, to enhance the power of models.
For \dg, ensemble learning exploits the relationship between multiple source domains by using specific network architecture designs and training strategies to improve generalization.
They assume that any sample can be regarded as an integrated sample of the multiple source domains, so the overall prediction result can be seen as the superposition of the multiple domain networks.

\citet{mancini2018best} proposed to use learnable weights for aggregating the predictions from different source specific classifiers, where a domain predictor is used to predict the probability that a sample belongs to each domain (weights). 
\citet{segu2020batch} maintained domain-dependent batch normalization (BN) statistics and BN parameters for different source domains while all the other parameters were shared. In inference, the final prediction was a linear combination of the domain-dependent models with the combination weights inferred by measuring the distances between the instance normalization statistics of the test sample and the accumulated population statistics of each domain. 
The work of \cite{d2018domain} proposed domain-specific layers of different source domains and learning the linear aggregation of these layers to represent a test sample.
\citet{zhou2020domain} proposed Domain Adaptive Ensemble Learning (DAEL), where a DAEL model is composed of a CNN feature extractor shared across domains and multiple domain-specific classifier heads. Each classifier is an expert to its own domain and a non-expert to others. DAEL aims to learn these experts collaboratively, by teaching the non-experts with the expert so as to encourage the ensemble to learn how to handle data from unseen domains.
There are also other works~\cite{wu2021collaborative,dubey2021adaptive}.
Ensemble learning remains a powerful tool for DG since ensemble allows more diversity of models and features.
However, one drawback of ensemble learning-based DG is maybe its computational resources as we need more space and computations for training and saving different models.



\subsubsection{Meta-learning-based DG}
The key idea of meta-learning is to learn a general model from multiple tasks by either optimization-based methods~\cite{Finn2017ModelAgnosticMF}, metric-based learning~\cite{snell2017prototypical}, or model-based methods~\cite{santoro2016meta}.
The idea of meta-learning has been exploited for \dg. 
They divide the data form multi-source domains into meta-train and meta-test sets to simulate domain shift.
Denote $\theta$ the model parameters to be learned, meta-learning can be formulated as:
\begin{equation}
    \label{eq-meta}
    \begin{split}
    \theta^\ast &= \operatorname{Learn}(\mathcal{S}_{mte}; \phi^\ast)\\ &= \operatorname{Learn}(\mathcal{S}_{mte}; \operatorname{MetaLearn}(\mathcal{S}_{mtrn})),
    \end{split}
\end{equation}
where $\phi^\ast = \operatorname{MetaLearn}(\mathcal{S}_{mtrn})$ denotes the meta-learned parameters from the meta-train set $\mathcal{S}_{mtrn}$ which is then used to learn the model parameters $\theta^\ast$ on the meta-test set $\mathcal{S}_{mte}$.
The two functions $\operatorname{Learn}(\cdot)$ and $\operatorname{MetaLearn}(\cdot)$ are to be designed and implemented by different meta-learning algorithms, which corresponds to a bi-level optimization problem.
The gradient update can be formulated as:
\begin{equation}
    \theta = \theta - \alpha \frac{\partial (\ell(\mathcal{S}_{mte};\theta) + \beta \ell(\mathcal{S}_{mtrn};\phi))}{\partial \theta},
\end{equation}
where $\eta$ and $\beta$ are learning rates for outer and inner loops, respectively.

\citet{Finn2017ModelAgnosticMF} proposed  Model-agnostic meta-learning (MAML). Inspired by MAML, \citet{li2018learning2} proposed MLDG (meta-learning for \dg) to use the meta-learning strategy for DG.
MLDG splits the data from the source domains into meta-train and meta-test to simulate the domain shift situation to learn general representations.
\citet{balaji2018metareg} proposed to learn a meta regularizer (MetaReg) for the classifier. \cite{li2019feature} proposed feature-critic training for the feature extractor by designing a meta optimizer.
\citet{Dou2019DomainGV} used the similar idea of MLDG and additionally introduced two complementary losses to explicitly regularize the semantic structure of feature space.
\citet{Du2020LearningTL} proposed an extended version of information bottleneck named Meta Variational Information Bottleneck (MetaVIB).
They regularize the Kullback–Leibler (KL) divergence between distributions of latent encoding of the samples that have the same category from different domains and learn to generate weights by using stochastic neural networks.
Recently, some works also adopted meta-learning for semi-supervised DG or discriminative DG~\cite{chen2020discriminative,sharifi2020domain,wang2020meta,zhao2020learning,zhao2021learning}.
Meta-learning is widely adopted in DG research and it can be incorporated into several paradigms such as disentanglement~\cite{bui2021exploiting}.
Meta-learning performs well on massive domains since meta-learning can seek transferable knowledge from multiple tasks.

\subsubsection{Gradient operation-based DG}

Other than meta-learning and ensemble learning, several recent works consider using gradient information to force the network to learn generalized representations.
\citet{huang2020self} proposed a self-challenging training algorithm that aims to learn general representations by manipulating gradients. 
They iteratively discarded the dominant features activated on the training data, and forced the network to activate remaining features that correlate with labels. In this way, network can be forced to learn from more bad cases which will improve generalization ability.
\citet{shi2021gradient} proposed a gradient-matching scheme, where their assumption is that the gradient direction of two domains should be the same to enhance common representation learning.
To this end, they proposed to maximize the gradient inner product (GIP) to align the gradient direction
across domains. With this operation, the network can find weights such that the input-output correspondence is as close as possible
across domains. GIP can be formulated as:
\begin{equation}
    \mathcal{L}=\mathcal{L}_{\text {cls }}\left(\mathcal{S}_{t rain} ; \theta\right)-\lambda \frac{2}{M(M-1)} \sum_{i,j}^{i \neq j} G_{i} \cdot G_{j},
\end{equation}
where $G_i$ and $G_j$ are gradient for two domains that can be calulated as $G=\mathbb{E} \frac{\partial \ell(x, y; \theta)}{\partial \theta}$.
The gradient-invariance was achieved by adding CORAL~\cite{sun2016deep} loss between gradients in \cite{rame2021fishr}, while \citet{tian2021neuron} maximized the neuron coverage of
DNN with gradient similarity regularization between the
original and the augmented samples.
Additionally, \citet{wang2021embracing} designed a knowledge distillation approach for based on gradient filtering.

\subsubsection{Distributionally robust optimization-based DG}

The goal of distributionally robust optimization (DRO)~\cite{rahimian2019distributionally} is to learn a model at worst-case distribution scenario to hope it can generalize well to the test data, which shares similar goal as DG.
To optimize the worst-case distribution scenario, \citet{Sagawa2019DistributionallyRN} proposed a GroupDRO algorithm that requires explicit group annotation of the samples.
Such annotation was later narrowed down to a small fraction of validation set in \cite{liu2021just}, where they formulated a two-stage weighting framework.
Other researchers reduced the variance of training domain risks by risk extrapolation (VRex)~\cite{krueger2021out} or reducing class-conditioned Wasserstein DRO~\cite{wang2021class}.
Recently, \citet{koh2020wilds} proposed the setting of subpopulation shift where they also applied DRO to this problem.
Particularly, \citet{du2021adarnn} proposed AdaRNN, a similar algorithm to the spirit of DRO that did not require explicit group annotation; instead, they learned the worst-case distribution scenario by solving an optimization problem.
To summarize, DRO focuses on the optimization process that can also be leveraged in DG research.

\subsubsection{Self-supervised learning-based DG}

Self-supervised learning (SSL) is a recently popular learning paradigm that builds self-supervised tasks from the large-scale unlabeled data~\cite{jing2020self}.
Inspired by this, \citet{carlucci2019domain} introduced a self-supervision task of solving jigsaw puzzles to learn generalized representations.
Apart from introducing new pretext tasks, contrastive learning is another popular paradigm of self-supervised learning, which was adopted in several recent works~\cite{kim2021selfreg,li2021domain,jeon2021feature}.
The core of contrastive learning is to perform unsupervised learning between positive and negative pairs.
Note that self-supervised learning is a general paradigm that can be applied to any existing DG methods, especially unsupervised DG where there are no labels in training domains~\cite{qi2021unsupervised}.
Another possible application of SSL-based DG is the pretraining of multi-domain data that trains powerful pretraining models while also handling domain shifts.
However, a possible limitation of SSL-based DG maybe its computational efficiency and requirement of computing resources.

\subsubsection{Other learning strategy for DG}
There are some other learning strategies for \dg.
For instance, metric learning was adopted in \cite{faraki2021cross} to explore better pair-wise distance for DG.
\citet{ryu2019generalized} used random forest to improve the generalization ability of convolutional neural networks (CNN). They sampled the triplets based on the probability mass function of the split results given by random forest, which is used for updating the CNN parameters by triplet loss.
Other works~\cite{wald2021calibration,gong2021confidence} adopted model calibration for DG, where they argued that the calibrated performance has a close relationship with OOD performance.
\citet{zhang2021can} followed the lottery ticket hyphothesis to design network substructures for DG, while \citet{narayanan2021shape} focused on the shape-invariant features.
Additionally, \citet{cha2021swad} observed the flat minima is important to DG and they designed a simple stochastic weight averaging densely method to find the flat minima.
Since DG is a general learning problem, there will be more works that uses other strategy in the future.

\section{Other Domain Generalization Research Areas}
\label{sec-research-area}
Most of the existing literature on \dg adopt the basic (traditional) definition of DG in Def.~\ref{def-dg}.
There is some existing literature that extends such setting to new scenarios to push the frontiers of DG (ref. \tablename~\ref{tb-dg-area}).
This section briefly discuss the existing DG research areas to give the readers a brief overview of this problem.

\begin{table*}[t!]
\centering
\caption{Existing research areas of domain generalization}
\label{tb-dg-area}
\begin{tabular}{lll}
\toprule
Setting &
  Definition &
  Reference \\ \midrule
Traditional domain generalization &
  Def.~\ref{def-dg} &
  Most of this paper \\ 
Single-source domain generalization &
  Set $M=1$ in Def.~\ref{def-dg} &
  \cite{duboudin2021encouraging,jia2020single,kim2021selfreg,li2021progressive,ouyang2021causality,peng2021out,qiao2020learning,duboudin2021encouraging,fan2021adversarially,wang2021learningtodiversify} \\ 
Semi-supervised domain generalization &
  $\mathcal{S}_{train}$ is partially labeled &
  \cite{zhou2021semi,lin2021semi} \\ 
Federated domain generalization &
  $\mathcal{S}_{train}$ cannot broadcast to the server &
  \cite{zhang2021federated,liu2021feddg,wu2021collaborative} \\ 
Open domain generalization &
  $\mathcal{Y}_{train} \ne \mathcal{Y}_{test}$ &
  \cite{shu2021open} \\ 
Unsupervised domain generalization &
  $\mathcal{S}_{train}$ is totally unlabeled &
  \cite{qi2021unsupervised} \\ \bottomrule
\end{tabular}%
\end{table*}

\subsection{Single-source Domain Generalization}

Setting $M=1$ in Def.~\ref{def-dg} gives single-source DG.
Compared to traditional DG ($M>1$), single-source DG becomes more challenging since there are less diversity in training domains.
Thus, the key to this problem is to generate novel domains using data generation techniques to increase the diversity and informativeness of training data.
Several methods designed different generation strategies~\cite{duboudin2021encouraging,jia2020single,kim2021selfreg,li2021progressive,ouyang2021causality,peng2021out,qiao2020learning,fan2021adversarially,wang2021learningtodiversify} for single-source DG in computer vision tasks.
A recent work~\cite{du2021adarnn} studied this setting in time series data where there is usually one unified dataset by using min-max optimization.
We expect more application areas can benefit from single-source DG.

\subsection{Semi-supervised Domain Generalization}

Compared to traditional DG, semi-supervised DG does not require the full labels of training domains.
It is common to apply existing semi-supervised learning algorithms such as FixMatch~\cite{sohn2020fixmatch} and FlexMatch~\cite{zhang2021flexmatch} to learn pseudo labels for the unlabeled samples.
For instance, two recent works adopted the consistency regularization in semi-supervised learning~\cite{zhou2021semi,lin2021semi} for semi-supervised DG.
It can be seen that this setting is more general than traditional DG and we expect there will be more works in this area.

\subsection{Federated Learning with Domain Generalization}
Privacy and security of machine learning is becoming increasingly critical~\citep{zhang2022remos}.
\citet{mahajan2021connection} firstly studied this problem and showed that if the features are stable, then the model is more robust to membership inference attack.
In federated DG~\cite{yang2019federated,yang2019federatedtist}, models do not access the raw training data; instead, they aggregate the parameters from different clients.
Under this circumstance, the key is to design better aggregation scheme through generalization techniques~\cite{zhang2021federated,liu2021feddg,wu2021collaborative}.
Federated DG is more important in healthcare~\cite{chen2020fedhealth}. On the other hand, decentralized training is another possible solution~\cite{kaissis2020secure}. However, similar privacy risks emerge when there is a need to update the model. Thus, we hope there could be more research.

\subsection{Other DG Settings}

There are also other settings in DG, such as open \dg~\cite{shu2021open} and unsupervised \dg~\cite{qi2021unsupervised}.
Open DG shares the similar setting of universal domain adaptation where the training and test label spaces are not the same.
Unsupervised DG assumes that all labels in the training domains are not accessible.
As the environment gets more general and challenging, there will be other DG research areas aiming at solving certain limitations.

\section{Applications}
\label{sec-app}

\begin{figure}[t!]
	\centering
	\resizebox{.5\textwidth}{!}{
	\begin{forest}
  for tree={
  grow=east,
  reversed=true,
  anchor=base west,
  parent anchor=east,
  child anchor=west,
  base=left,
  font=\small,
  rectangle,
  draw,
  rounded corners,align=left,
  minimum width=2.5em,
  inner xsep=4pt,
  inner ysep=1pt,
  },
  where level=1{text width=5em,fill=blue!10}{},
  where level=2{text width=5em,font=\footnotesize,fill=pink!30}{},
  where level=3{font=\footnotesize,yshift=0.26pt,fill=yellow!20}{},
  [Domain\\generalization\\applications,fill=green!20
    [Computer\\vision,text width=4.5em
        [Image classification,text width=7.5em
            [e.g. DRPC \cite{yue2019domain}/ CSG \cite{liu2021learning}
            ]
        ]
        [Semantic segmentation,text width=9em
          [e.g. DLOW \cite{gong2019dlow}/ \cite{li2021semantic}
          ]
        ]
        [Action recognition,text width=8em
            [e.g. Epi-FCR \cite{li2019episodic}/ \cite{li2017domain}
            ]
        ]
        [Person Re-ID,text width=5.5em
            [e.g. DomainMix \cite{wang2020domainmix}/ FAR \cite{jin2020feature}
            ]
        ]
    ]
    [Natural\\language\\processing,text width=5em
        [Sentiment classification,text width=9em
            [e.g. MetaReg \cite{balaji2018metareg}/ \cite{wang2020unseen}
            ]
        ]
        [Semantic parsing,text width=7em
         [e.g. \cite{wang2020meta}
         ]
        ]
        [Web page classification,text width=9em
            [e.g. FSUS \cite{garg2020learn}
            ]
        ]
    ]
    [Reinforcement\\learning,text width=6em
        [Robot control,text width=5em
            [e.g. MLDG \cite{li2018learning2}/ MixStyle \cite{zhoudomain}
            ]
        ]
    ]
    [Medical\\analysis,text width=5em
        [Tissue segmentation,text width=8em
            [e.g. MASF \cite{Dou2019DomainGV}
            ]
        ]
        [Parkinson's disease,text width=8em
            [e.g. DICA \cite{muandet2013domain}
            ]
        ]
        [Chest X-ray recognition,text width=9em
            [e.g. MatchDG \cite{Mahajan2020DomainGU}/ \cite{li2020domain}
            ]
        ]
        [EEG-based seizure detection,text width=11em
            [e.g. \cite{ayodele2020supervised}
            ]
        ]
    ]
    [Other\\applications,text width=5em
        [Human activity recognition,text width=10.5em
            [e.g. GILE \cite{qian2021latent}/ LAG \cite{lu2022local}
            ]
        ]
        [Fault diagnosis,text width=6em
            [e.g. DSDGN \cite{liao2020deep}/ DDGFD \cite{zheng2020deep}/ \cite{li2020domain}
            ]
        ]
        [Time series forecasting,text width=9em
            [e.g. AdaRNN \cite{du2021adarnn}
            ]
        ]
        [Speech utterance recognition,text width=10.8em
            [e.g. CrossGrad \cite{shankar2018generalizing}/CSD \cite{piratla2020efficient}
            ]
        ]
    ]
]
\end{forest}

        
	}
	\caption{Several applications of \dg.}
	\label{fig-app}
\end{figure}
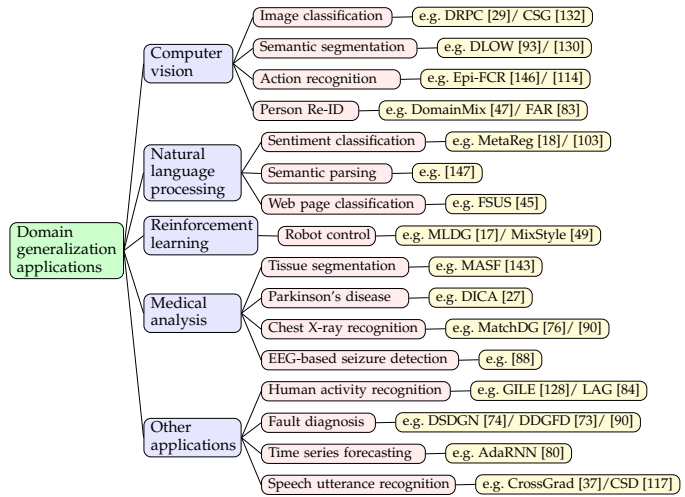

In this section, we discuss the popular tasks/applications for \dg (ref. \figurename~\ref{fig-app}).


High generalization ability is desired in various vision tasks. Many works investigate DG on classification. Some works also study DG for semantic segmentation~\cite{gong2019dlow}, action recognition~\cite{li2017domain,li2019episodic}, face anti-spoofing~\cite{shao2019multi}, person Re-ID~\cite{wang2020domainmix,jin2020feature}, street view recognition~\cite{qiao2020learning}, video understanding~\cite{niu2015multi}, and image compression~\cite{zhang2021out}.
Medical analysis~\cite{hu2021domain} is one of the important application areas for DG due to its nature of data scarcity and existence of domain gaps, with the tasks of tissue segmentation~\cite{Dou2019DomainGV}, Parkinson's disease recognition~\cite{muandet2013domain}, activity recognition~\cite{erfani2016robust}, chest X-ray recognition~\cite{Mahajan2020DomainGU,li2020domain}, and EEG-based seizure detection~\cite{ayodele2020supervised}. 

Apart from those areas, DG is also useful in reinforcement learning of robot control~\cite{zhoudomain,li2018learning2} to generalize to an unseen environment.
Some work used DG to recognize speech utterance~\cite{shankar2018generalizing,piratla2020efficient}, fault diagnosis~\cite{li2020domain,liao2020deep,zheng2020deep}, physics~\cite{chen2020study}, brain-computer interface~\cite{han2020domain}.

In natural language processing, it is also common that training data comes from different domains with different distributions and DG techniques are helpful. 
Some work used \dg for sentiment classification on the Amazon Review dataset~\cite{wang2020unseen,balaji2018metareg}.
Others used DG for semantic parsing~\cite{wang2020meta}, web page classification~\cite{garg2020learn}. For instance, if we are given natural language data from multiple domains and want to learn a generalized model that predicts well on any new domain, we can use \dg to acquire domain-invariant representations.

Moreover, DG techniques favor broad prospects in some applications, such as financial analysis, weather prediction, and logistics.
For instance, \citet{du2021adarnn} tried to adopt DG to time series modeling. They first propose the temporal covariate shift problem that widely exists in time series data, then, they proposed an RNN-based model to solve this problem to align the hidden representations between any pair of training data that are from different domains.
Their algorithm, the so-called AdaRNN, was applied to stock price prediction, weather prediction, and electric power consumption.
Another example is \cite{qian2021latent}, where they applied domain generalization to sensor-based human activity recognition.
In their application, the activity data from different persons are from different distributions, resulting in severe model collapse when applied to new users.
To resolve such problem, they developed a variational autoencoder-based network to learn the domain-invariant and domain-specific modules, thus achieving the disentanglement.

In the future, we hope there can be more DG applications in other areas to tackle with the distributional shift that widely exists in different applications.
Another important problem is the evaluation of DG algorithms without accessing the test distribution in reality. While we can use the test data for evaluation in research, we simply cannot do it for real applications. In this case, one possible approach would be performing meta-train and meta-test split for the original data for multiple times. In each time, one split can be regarded as the unseen test data while the other as the training data. We can call it the meta-cross-validation for DG in reality. At the same time, we also hope there could be more evaluation metrics. For more evaluation in research, please refer to the next section.

\begin{table*}[t!]
  \centering
  \caption{Eighteen popular datasets for \dg. The last ten datasets are from WILDS~\cite{koh2020wilds}.}
  \label{tb-dataset}%
    \begin{tabular}{lrrrll}
    \toprule
    Dataset & \#Domain & \#Class & \#Sample & Description & Reference \\ \hline
    Office-Caltech & 4 & 10 & 2,533  & Caltech, Amazon, Webcam, DSLR & \cite{saenko2010adapting} \\
    Office-31 & 3 & 31 & 4,110  & Amazon, Webcam, DSLR & \cite{saenko2010adapting} \\
    PACS  & 4     & 7     & 9,991  & Art, Cartoon, Photos, Sketches & \cite{li2017deeper} \\
    VLCS  & 4     & 5     & 10,729  & Caltech101, LabelMe, SUN09, VOC2007 & \cite{fang2013unbiased} \\
    Office-Home & 4     & 65    & 15,588  & Art, Clipart, Product, Real & \cite{Venkateswara2017DeepHN} \\
    Terra Incognita & 4     & 10    & 24,788  & Wild animal images taken at locations L100, L38, L43, L46 & \cite{Beery2018RecognitionIT} \\
    Rotated MNIST & 6     & 10    & 70,000  & Digits rotated from $0^\circ$ to $90^\circ$ with an interval of $15^\circ$ & \cite{Ghifary2015DomainGF} \\
    DomainNet & 6     & 345   & 586,575 & Clipart, Infograph, Painting, Quickdraw, Real, Sketch & \cite{Peng2019MomentMF} \\
    iWildCam2020-wilds  & 323     & 182                                                     & 203,029 & Species classification across different   camera traps          & \cite{beery2020iwildcam}                         \\
    Camelyon17-wilds    & 5       & 2                                                       & 45,000  & Tumor identification across five different hospitals            & \cite{bandi2018detection}                        \\
    RxRx1-wilds         & 51      & 1,139                                                   & 84,898  & Genetic perturbation classification across experimental batches & \cite{taylor2019rxrx1}                           \\
    OGB-MolPCBA         & 120,084 & 128                                                     & 400,000 & Molecular property prediction across different scaffolds        & \cite{hu2020open}                                \\
    GlobalWheat-wilds   & 47      & bounding boxes     & 6,515   & Wheat head detection across regions of the world                & \cite{david2020global}          \\
    CivilComments-wilds & -       & 2                                                       & 450,000 & Toxicity classification across demographic identities           & \cite{borkan2019nuanced}                         \\
    FMoW-wilds          & 80      & 62                                                      & 118,886 & Land use classification across different regions and years      & \cite{christie2018functional}                    \\
    PovertyMap-wilds    & 46      & real value & 19,669  & Poverty mapping across different countries                      & \cite{yeh2020using}                              \\
    Amazon-wilds        & 3920    & 5                                                       & 539,502 & Sentiment classification across different users                 & \cite{ni2019justifying}                          \\
    Py150-wilds         & 8,421   & next token                                              & 150,000 & Code completion across different codebases                      & \cite{lu2021codexglue, raychev2016probabilistic}\\
    \bottomrule
    \end{tabular}
    
\end{table*}%

\section{Datasets, Evaluation, and Benchmark}
\label{sec-dataset}

In this section, we summarize the existing common datasets and model selection strategies for \dg.
Then, we introduce the codebase, DeepDG, and demonstrate some observations from experiment conducted via it.

\subsection{Datasets}
\tablename~\ref{tb-dataset} offers an overview of several popular datasets.
Among them, PACS~\cite{li2017deeper}, VLCS~\cite{fang2013unbiased}, and Office-Home~\cite{Venkateswara2017DeepHN} are three most popular datasets.
For large-scale evaluation, DomainNet~\cite{Peng2019MomentMF} and Wilds~\cite{koh2020wilds} (\emph{i.e.,} a collection of datasets in \tablename~\ref{tb-dataset} with `-wilds') are becoming popular.

Besides the datasets mentioned above, there exists some other datasets for \dg with different tasks.
The Graft-versus-host disease (GvHD) dataset~\cite{brinkman2007high} is also popular and is used to test several methods~\cite{muandet2013domain,blanchard2011generalizing} for the flow cytometry problem.
This dataset was collected from 30 patients with sample sizes ranging from 1,000 to 10,000. This is a time series classification dataset. 
Some works~\cite{gong2019dlow,yue2019domain,jin2021style} applied \dg to semantic segmentation, where CityScape~\cite{cordts2016cityscapes} and GTA5~\cite{richter2016playing} datasets were adopted as benchmark datasets.
Some works applied DG to object detection, using the datasets of Cityscapes~\cite{cordts2016cityscapes}, GTA5~\cite{richter2016playing}, Synthia~\cite{ros2016synthia} for investigation~\cite{jin2021style}.
Some other works used public datasets or RandPerson~\cite{wang2020surpassing} for person re-identification \cite{jia2019frustratingly,song2019generalizable,jin2020style}. Some works~\cite{li2018learning2,zhoudomain} used the OpenAI Gym~\cite{brockman2016openai} as the testbed to evaluate the performance of algorithms in reinforcement learning problems such as Cart-Pole and mountain car.

In addition to these widely used datasets, there are also other datasets used in existing literature.
The Parkinson’s telemonitoring dataset~\cite{little2008suitability} is popular for predicting the clinician’s motor and total UPDRS scoring of Parkinson’s disease symptoms from voice measures.
Some methods~\cite{muandet2013domain,blanchard2017domain,grubinger2015domain} used the data from several people as the training domains to learn models that generalize to unseen subjects.



It is worth noting that the datasets of domain generalization have some overlaps with domain adaptation.
For instance, Office-31, Office-Caltech, Office-Home, and DomainNet are also widely used benchmarks for \da.
Therefore, most domain adaptation datasets can be used for \dg benchmark in addition to those we discussed here.
For example, Amazon Review dataset~\cite{blitzer2006domain} is widely used in \da. It has four different domains on product review (DVDs, Kitchen appliance, Electronics and Books), which can also be used for \dg.

\subsection{Evaluation}
To test \dg algorithm on a test domain, three strategies are proposed~\cite{gulrajani2021search}, namely, \emph{Test-domain validation set}, \emph{Leave-one-domain-out cross-validation}, and \emph{Training-domain validation set}. 
Test-domain validation set utilizes parts of the target domain as validations.
Although it can obtain the best performance in most circumstances for that validation and testing share the same distribution, there is often no access to targets when training, which means it cannot be adopted in real applications.
Leave-one-domain-out is another strategy to choose the final model when training data contains multiple sources.
It leaves one training source as the validation while treating the others as the training part.
Obviously, when only a single source exists in the training data, it is no longer applicable.
In addition, due to different distributions among sources and targets, final results rely heavily on the selections of validation, which makes final results unstable.
The most common strategy for \dg is Training-domain validation set which is used in most existing work.
In this strategy, each source is split into two parts, the training part and the validation part.
All training parts are combined for training while all validation parts are combined for selecting the best model.
Since there still exists divergences between the combined validation and the real unseen targets, this simple and most popular strategy cannot achieve the best performance for some time.

We need to mention that there may exist other evaluation protocols for DG such as \citep{ye2021towards} since designing effective evaluation protocols is often consistent with the OOD performance.
Currently, most of the works adopted the train-domain validation strategy which may not always generate good performance since the distribution of validation set is not the same as the new training data. On the other hand, using accuracy alone may not be sufficient to valid the model performance. We are looking forward to new evaluation metrics that can truly reflect the properties test distributions as much as possible in order to obtain better results.\looseness=-1

\subsection{Benchmark}
To test the performance of DG algorithms in a unified codebase, in this paper, we develop a new codebase for DG, named \emph{DeepDG}~\cite{deepdg,transferlearning.xyz}.
Compared to the existing DomainBed~\cite{gulrajani2021search}, DeepDG simplifies the data loading and model selection process, while also makes it possible to run all experiments in a single machine.
DeepDG splits the whole process into a data preparation part, a model part, a core algorithm part, a program entry, and some other auxiliary functions.
Each part can be freely modified by users without affecting other parts.
Users can add their own algorithms or datasets to DeepDG and compare with some state-of-the-art methods fairly.
The current public version of DeepDG is only for image classification and we offer supports for Office-31, PACS, VLCS, and Office-Home datasets.
Currently, nine state-of-the-art methods are implemented under the same environment, and it covers all three groups, including Data manipulation (Mixup~\cite{zhang2018mixup}), Representation learning (DDC~\cite{tzeng2014deep}, DANN~\cite{ganin2015unsupervised}, CORAL~\cite{sun2016return}), and Learning strategy (MLDG~\cite{li2018learning2}, RSC~\cite{huang2020self}, GroupDRO~\cite{sagawa2019distributionally}, ANDMask~\cite{parascandolo2020learning}).

We conduct some experiments on the two most popular image classification datasets, PACS and Office-Home, with DeepDG and \tablename~\ref{tab:my-table-deepdg} shows the results.
ResNet-18 is used as the base feature network. 
Training-domain validation set is used for selecting final models, and $20\%$ of sources are for validation while the others are for training.
From \tablename~\ref{tab:my-table-deepdg}, we observe more insightful conclusions.
(1) The baseline method, ERM, has achieved acceptable results on both datasets. Some methods, such as DANN and ANDMask, even have worse performance.
(2) Simple data augmentation method, Mixup, cannot obtain remarkable results. 
(3) CORAL has slight improvements on both datasets compared to ERM, which is consistent with results offered by DomainBed~\cite{gulrajani2021search}.
(4) RSC, a learning strategy, achieves the best performance on both datasets, but the improvements are unremarkable compared to ERM.
The results indicate the benefits of domain generalization in different tasks.

\begin{table}[!t]
\centering
\caption{Benchmark results for PACS and Office-Home with DeepDG.}
\label{tab:my-table-deepdg}
\resizebox{.5\textwidth}{!}{
\begin{tabular}{l|ccccc|ccccc}
\toprule
Dataset  & \multicolumn{5}{c|}{PACS}              & \multicolumn{5}{c}{Office-Home}       \\
Method   & A     & C     & P     & S     & AVG   & A     & C     & P     & R     & AVG   \\
\midrule
ERM      & 77.0 & 74.5 & 95.5 & 77.8 & 81.2 & 58.6 & 47.9 & 72.2 & 73.0 & 62.9 \\
DANN~\cite{Ganin2016DomainAdversarialTO}     & 78.7 & 75.3 & 94.0 & 77.8 & 81.4 & 57.7 & 44.4 & 71.9 & 72.5  & 61.6 \\
CORAL~\cite{sun2016deep}    & 77.7 & 77.0 & 92.6 & 80.5 & 82.0 & 58.7 & 48.7 & 72.3 & 73.6 & 63.3 \\
Mixup~\cite{zhang2018mixup}    & 79.1 & 73.4 & 94.4 & 76.7 & 80.9 & 55.7 & 47.9 & 71.9 & 72.8 & 62.1 \\
RSC~\cite{huang2020self}      & 79.7 & 76.1 & 95.6 & 76.6 & 82.0 & 58.9 & 49.2 & 72.5 & 74.2 & 63.7  \\
GroupDRO~\cite{Sagawa2019DistributionallyRN} & 76.0 & 76.0 & 91.2 & 79.0 & 80.6 & 57.6  & 48.7 & 71.5 & 73.1 & 62.7 \\
ANDMask~\cite{parascandolo2020learning}  & 76.2 & 73.8 & 91.6 & 78.0 & 79.9 & 56.7 & 45.9 & 70.6 & 73.2 & 61.6 \\ \bottomrule
\end{tabular}}
\vspace{-.2in}
\end{table}

\section{Discussion}
\label{sec-diss}

In this section, we summarize existing methods and then present several
challenges for future. 

\subsection{Summary of Existing Literature}
The quantity and diversity of training data are critical to a model's generalization ability.
Many methods aim to enrich the training data with the data manipulation methods to achieve good performance.
However, one issue of the data manipulation methods is that there is a lack of theoretical guarantee of the unbound risk of generalization.
Therefore, it is important to develop theories for the manipulation-based methods which could further guide the data generation designs without violating ethical standards.

Compared to data manipulation, representation learning has theoretical support in general \citep{ben2007analysis,blanchard2011generalizing,albuquerque2019adversarial}.
Kernel-based methods are widely used in traditional methods while deep learning-based methods play a leading role in recent years.
While domain adversarial training often achieves better performance in domain adaptation, in DG, we did not see significant results improvements from these adversarial methods. We think this is probably because the task is relatively easy.
For the explicit distribution matching, more and more works tend to match the joint distributions rather than just match the marginal~\citep{blanchard2011generalizing,Li2018DomainGW} or conditional~\citep{li2018domain} distributions.
Thus, it is more feasible to perform dynamic distribution matching~\citep{wang2018visual,wang2020transfer}.
Both disentanglement and IRM methods have good motivations for generalization, while more efficient training strategy can be developed.
There are several studies \cite{shui2021benefits} that pointed out merely learning domain-invariant features are insufficient and representation smoothness should also be considered.

For learning strategy, there is a trend that many works used meta-learning for DG, where it requires to design better optimization strategies to utilize the rich information of different domains.
In addition to deep networks, there are also some work~\citep{ryu2019generalized} that used random forest for DG, and we hope more diverse methods will come.

\subsection{Future Research Challenges}



\subsubsection{Continuous \dg}
For many real applications, a system consumes streaming data with non-stationary statistics. 
In this case, it is of great importance to perform continuous \dg that efficiently updates DG models to overcome catastrophic forgetting and adapts to new data.
While there are some \da methods focusing on continuous learning~\cite{wang2020continuously}, there are only very few investigations on continuous DG \cite{li2020sequential} whenever this is favorable in real scenarios.

\subsubsection{Domain generalization to novel categories}
The existing DG algorithms usually assume the label space for different domains are the same.
A more practical and general setting is to support the generalization on new categories, \textit{i.e.}, both domain and task generalization.
This is conceptually similar to the goal of meta-learning and zero-shot learning.
Some work~\cite{maniyar2020zero,mancini2020towards} proposed zero-shot DG and we expect more work to come in this area. There are some prior work \cite{shu2021open,zhu2021crossmatch} that tried to use the boundary-based learning paradigms or consistency regularization to solve this problem, which are good approaches that future work might build methods upon them.

\subsubsection{Interpretable domain generalization}
Disentanglement-based DG methods decompose a feature into domain-invariant/shared and domain-specific parts, which provide some interpretation to DG.
For other categories of methods, there is still a lack of deep understanding of the semantics or characteristics of learned features in DG models.
For example, how to relate the results of the approach with the input
feature space. How close are current methods to provide this level of interpretability?
Causality~\citep{liu2021learning} may be one promising tool to understand \dg networks and provide interpretations.

\subsubsection{Large-scale pre-training/self-learning and DG}
In recent years, we have witnessed the rapid development of large-scale pre-training/self-learning, such as BERT~\cite{devlin2018bert}, GPT-3~\cite{brown2020language}, and Wav2vec~\cite{baevski2020wav2vec}.
Pre-training on large-scale dataset and then finetuning the model to downstream tasks could improve its performance, where pre-training is beneficial to learn general representations. Therefore, how to design useful and efficient DG methods to help large-scale pre-training/self-learning is worth investigating. 

\subsubsection{Test-time Generalization}

While DG focuses on the training phase, we can also request test-time generalization in inference phase.
This further bridges \da and \dg since we can also use the inference unlabeled data for adaptation.
Very few recent works~\cite{iwasawa2021test,pandey2021domain} paid attention to this setting.
Compared to traditional DG, test-time generalization will allow more flexibility in inference time, while it requires less computation and more efficiency as there are often limited resources in inference-end devices.

\subsubsection{Performance evaluation for DG}
The recent work~\cite{gulrajani2021search} pointed out that on several datasets, the performance of some DG methods is almost the same as the baseline (i.e., ERM).
We do not take it as the full evidence that DG is not useful in real applications. Instead, we argue that this might be due to the inappropriate evaluation schemes in use today, or the domain gaps not being so large.
In more realistic situations such as person ReID where there are obvious domain gaps~\cite{jin2021style}, the improvement of DG is dramatic.
Therefore, we stay positive about the value of DG and hope researchers can also find more suitable settings and datasets for the study.


\section{Conclusion}
\label{sec-con}
Generalization has always been an important research topic in machine learning research. In this paper, we review the \dg areas by providing in-depth analysis of theories, existing methods, datasets, benchmarks, and applications.
Then, we thoroughly analyze the methods.
Based on our analysis, we provide several potential research challenges that could be the directions of future research.
We hope that this survey can provide useful insights to researchers and inspire more progress in the future.

\section*{Acknowledgement}

This work is supported in part by NSFC (No. 61972383), NSF under grants III-1763325, III-1909323,  III-2106758, and SaTC-1930941. 



\bibliographystyle{IEEEtranN}
{\footnotesize
\bibliography{ijcai21}
}


\end{document}